\documentclass[journal]{IEEEtran}
\usepackage{times}
\usepackage{epsfig}
\usepackage{graphicx}
\usepackage{amsmath}
\usepackage{multirow}
\usepackage{amssymb}
\usepackage{subfigure}
\usepackage{mathrsfs}
\usepackage[ruled]{algorithm2e}
\usepackage{cite}
\usepackage[justification=centering]{caption}
\ifCLASSINFOpdf
\else
\fi
\begin{document}
%
% paper title
% Titles are generally capitalized except for words such as a, an, and, as,
% at, but, by, for, in, nor, of, on, or, the, to and up, which are usually
% not capitalized unless they are the first or last word of the title.
% Linebreaks \\ can be used within to get better formatting as desired.
% Do not put math or special symbols in the title.
\title{Shared Predictive Cross-Modal Deep Quantization}
%
%
% author names and IEEE memberships
% note positions of commas and nonbreaking spaces ( ~ ) LaTeX will not break
% a structure at a ~ so this keeps an author's name from being broken across
% two lines.
% use \thanks{} to gain access to the first footnote area
% a separate \thanks must be used for each paragraph as LaTeX2e's \thanks
% was not built to handle multiple paragraphs
%

\author{Erkun~Yang,
        Cheng~Deng,~\IEEEmembership{Member,~IEEE},
        Chao~Li,
        Wei~Liu,
        Jie~Li,
        Dacheng~Tao,~\IEEEmembership{Fellow,~IEEE}% <-this % stops a space
\thanks{E. Yang, C. Deng, C. Li and J. Li are with the School of Electronic Engineering, Xidian University, Xi'an 710071, China (e-mail: ekyang@stu.xidian.edu.cn; chdeng.xd@gmail.com; li\_chao@stu.xidian.edu.cn; leejie@mail.xidian.edu.cn).}% <-this % stops a space
\thanks{W. Liu is with Tencent AI Lab, Shenzhen, China e-mail: (wliu@ee.columbia.edu).}% <-this % stops a space
\thanks{D. Tao is with School of Information Technologies, the University of Sydney, NSW 2007, Australia (email: dacheng.tao@sydney.edu.au).}
\thanks{\textcopyright 2018 IEEE. Personal use of this material is permitted. Permission from IEEE must be obtained for all other uses, in any current or future media, including reprinting/republishing this material for advertising or promotional purposes, creating new collective works, for resale or redistribution to servers or lists, or reuse of any copyrighted component of this work in other works.}}% <-this % stops a space
%\thanks{Manuscript received April 19, 2005; revised August 26, 2015.}}

% note the % following the last \IEEEmembership and also \thanks -
% these prevent an unwanted space from occurring between the last author name
% and the end of the author line. i.e., if you had this:
%
% \author{....lastname \thanks{...} \thanks{...} }
%                     ^------------^------------^----Do not want these spaces!
%
% a space would be appended to the last name and could cause every name on that
% line to be shifted left slightly. This is one of those "LaTeX things". For
% instance, "\textbf{A} \textbf{B}" will typeset as "A B" not "AB". To get
% "AB" then you have to do: "\textbf{A}\textbf{B}"
% \thanks is no different in this regard, so shield the last } of each \thanks
% that ends a line with a % and do not let a space in before the next \thanks.
% Spaces after \IEEEmembership other than the last one are OK (and needed) as
% you are supposed to have spaces between the names. For what it is worth,
% this is a minor point as most people would not even notice if the said evil
% space somehow managed to creep in.

% The paper headers
\markboth{IEEE Transaction on Neural Networks and Learning Systems}%
{Shell \MakeLowercase{\textit{et al.}}: Bare Demo of IEEEtran.cls for IEEE Journals}

% The only time the second header will appear is for the odd numbered pages
% after the title page when using the twoside option.
%
% *** Note that you probably will NOT want to include the author's ***
% *** name in the headers of peer review papers.                   ***
% You can use \ifCLASSOPTIONpeerreview for conditional compilation here if
% you desire.

% If you want to put a publisher's ID mark on the page you can do it like
% this:
%\IEEEpubid{0000--0000/00\$00.00~\copyright~2015 IEEE}
% Remember, if you use this you must call \IEEEpubidadjcol in the second
% column for its text to clear the IEEEpubid mark.

% use for special paper notices
%\IEEEspecialpapernotice{(Invited Paper)}

% make the title area
\maketitle

% As a general rule, do not put math, special symbols or citations
% in the abstract or keywords.
\begin{abstract}
With explosive growth of data volume and
ever-increasing diversity of data modalities, cross-modal similarity search, which conducts nearest neighbor search across different modalities, has been attracting increasing interest.
This paper presents a deep compact code learning solution for efficient cross-modal similarity search.
Many recent studies have proven that quantization based approaches perform generally better than hashing based
approaches on single-modal similarity search. In this work, we propose a deep  quantization approach, which
is among the early attempts of leveraging deep neural networks into quantization based cross-modal similarity
search. Our approach, dubbed shared predictive deep quantization (SPDQ), explicitly formulates a shared subspace
across different modalities and two private subspaces for individual modalities, representations in the shared subspace and the private subspaces are learned simultaneously by embedding them to a reproducing kernel Hilbert space where the mean embedding of different modality distributions can be explicitly compared. Additionally, in the shared subspace, a quantizer is learned to produce the semantics
preserving compact codes with the help of label alignment. Thanks to this novel network architecture in
cooperation with supervised quantization training, SPDQ can preserve intra- and inter-modal similarities
as much as possible and greatly reduce quantization error. Experiments on two popular benchmarks corroborate
that our approach outperforms state-of-the-art methods.
\end{abstract}

% Note that keywords are not normally used for peerreview papers.

% For peer review papers, you can put extra information on the cover
% page as needed:
% \ifCLASSOPTIONpeerreview
% \begin{center} \bfseries EDICS Category: 3-BBND \end{center}
% \fi
%
% For peerreview papers, this IEEEtran command inserts a page break and
% creates the second title. It will be ignored for other modes.
\IEEEpeerreviewmaketitle

\section{Introduction}
% The very first letter is a 2 line initial drop letter followed
% by the rest of the first word in caps.
%
% form to use if the first word consists of a single letter:
% \IEEEPARstart{A}{demo} file is ....
%
% form to use if you need the single drop letter followed by
% normal text (unknown if ever used by the IEEE):
% \IEEEPARstart{A}{}demo file is ....
%
% Some journals put the first two words in caps:
% \IEEEPARstart{T}{his demo} file is ....
%
% Here we have the typical use of a "T" for an initial drop letter
% and "HIS" in caps to complete the first word.
\IEEEPARstart{S}{imilarity} search is a fundamental subject in numerous computer vision applications, such as image and video retrieval~\cite{liu2012supervised,liu2017reversed,gui2017fast,wang2016learning,deng2013visual,chang2014factorized,liu2011hashing,liu2016sequential,gui2016supervised,zhao2015deep,ding2016defense,yu2012visual,qi2012exploring,you2017learning,liu2017boosting}, image classification~\cite{kulis2009kernelized}, object recognition~\cite{uijlings2013selective}, \textit{etc}. During the last decade, the amount of heterogeneous multimedia
data continues to grow at an astonishing speed, and multimedia data on the Internet usually exist in different media types and come from different data sources, e.g., video-tag pairs from YouTube, image-short text pairs from Facebook, and text-image pairs from news website. When we search a topic, it is expected to retrieve a ranked list containing data in various media types, which can give us a comprehensive description for the query topic.
So, efficient cross-modal similarity search becomes increasingly important
and has also been found far more challenging.

Actually, cross-modal retrieval has been widely studied in recent years~\cite{zhou2014latent, deng2016discriminative,cao2016deep,jiang2016deep,liu2015multi,liu2016query-adaptive,luo2013multiview,duan2012domain,qi2012clustering}.
In contrast to single-modal search scenario, heterogeneous data in cross-modal retrieval usually reside in different feature spaces, and how to exploit and build the correlation between heterogeneous modalities to preserve both intra- and inter-modal data similarities is a crucial issue. In order to eliminate the diversity between different modality features, recent studies are concentrated on mapping heterogeneous data into a common latent subspace so that the learned features in this subspace can be directly compared. However, due to high storage cost and low query efficiency, these methods can not deal with large-scale multi-modal data.

To tackle the efficiency and storage challenges, we study compact coding, a promising solution approaching cross-modal similarity search, especially focusing on a
common real-world cross-modal search scenario: image-to-text search. Compact coding methods transform high-dimensional
data points to indexable short binary codes, with which similarity search can be executed very efficiently. Most research
efforts have been made on cross-modal similarity search with typical solutions including hashing~\cite{bronstein2010data,liu2012compact,deng2015adaptive,deng2015large,ji2014query,liu2017scalable,kumar2011learning,liu2016query,liu2016joint,liu2016multilinear,yang2017pairwise,zhang2011composite,irie2015alternating,liu2014discrete,liu2014collaborative,lin2015semantics,yu2016binary,li2017linear}
and quantization~\cite{irie2015alternating,long2016composite,wu2015quantized}. Hashing based methods are usually proposed by mapping
original heterogeneous high-dimensional data into a common low-dimensional Hamming space and representing the original data by using a
set of compact binary codes. Quantization based methods, rather than using binary codes, usually approximate the original data by concatenating or adding
a set of learned quantizers. It has been proven that quantization enjoys
a more powerful representation ability than hashing, thanks to the more accuracy distance approximation~\cite{zhang2014composite,yu2010quantization,he2013k,zhang2015sparse}.
However, previous cross-modal hashing or quantization methods relying on shallow learning architectures cannot effectively
exploit the intrinsic relationships among different modalities.

Recently, deep hashing methods for cross-modal search~\cite{masci2014multimodal,jiang2016deep} emerge and yield attractive results on a number of benchmarks. Performance of those methods largely depends on whether those deep hashing models can effectively capture nonlinear correlations between different modalities. To further reduce
quantization error, deep quantization method~\cite{cao2017collective} is proposed for cross-modal similarity search,
where a deep representation and a quantizer for each modality are jointly learned with an end-to-end architecture. Unfortunately,
a common shortcoming of such deep methods, whether hashing or quantization, is that they individually construct two networks
for different modalities and then learn their corresponding deep representations, which isolates the relationships among different
modalities. Moreover, current quantization strategies do not consider the impact of semantic information (such as labels) on
the quality of quantization and similarity preserving.

\begin{figure}[!ht]
\centering
\includegraphics[width=0.4\textwidth]{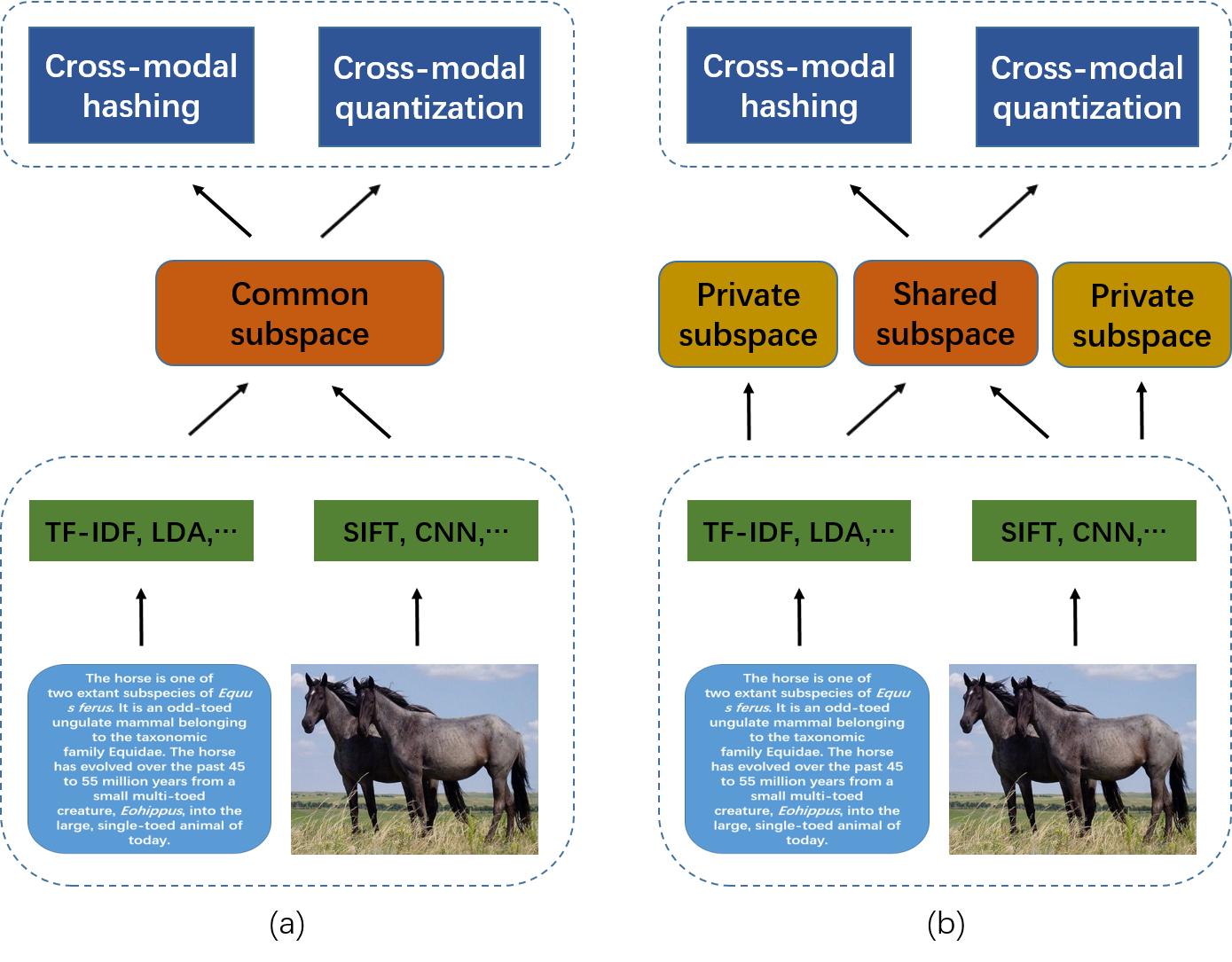}
\caption{The difference between traditional common subspace learning methods and the proposed SPDQ method: (a) traditional common subspace learning methods, (b) the proposed SPDQ method.}
\label{fig: 1}
\end{figure}

In this paper, we propose a novel quantization approach for cross-modal similarity search, dubbed Shared Predictive Deep
Quantization (SPDQ), aiming at adequately exploiting the intrinsic correlations among multiple modalities and learning
compact codes of higher quality in a joint deep network architecture. Specifically, we first adopt deep neural networks
to construct a shared subspace and two private subspaces respectively. The private subspaces are used to capture modality
specific properties, while the shared subspace is used to capture the representations shared by multiple modalities. Fig. ~\ref{fig: 1} illustrates the difference between traditional common subspace learning methods and the proposed method. Compared with traditional common subspace learning methods~\cite{yu2015hybrid,yu2016hybrid}, by
finding a shared subspace that is independent of the private subspaces, our proposed model can capture intrinsic semantic information shared between multi-modal data more efficiently. Actually, representations in shared subspace and private subspaces are learned
simultaneously by embedding them to a reproducing kernel
Hilbert space where the mean embedding of different modality
distributions can be explicitly compared. Moreover,
we quantize the representations to produce compact codes in the shared subspace, where label alignment is introduced to
enforce semantic similarities. In doing so, data points in the same class are encouraged to have the same representation,
therefore greatly reducing intra-class dissimilarity, then both intra- and inter-modal similarities are well preserved. Compared with existing
works, main contributions of the proposed SPDQ method are summarized as follows:
\begin{itemize}
\item By explicitly learning a shared subspace and two private subspaces, our method can extract the correlated information between different modalities more efficiently.
\item Representations from different classes are transformed and aligned using label information, which can greatly reduce the intra-class diversity, and both intra-modal and inter-modal semantic similarities are well preserved.
\item By integrating shared and private subspace learning and representation quantization in an end-to-end mechanism, our method can jointly optimize each part and generate more discriminative representations which are suitable for quantization.
\item Experimental results on two popular cross-modal datasets show that our approach significantly outperforms state-of-the-arts in terms of search accuracy and search efficiency.
\end{itemize}

The rest of this paper is organized as follows. We review the relevant literature in Section~\ref{sec:related work}. We present our novel deep quantization approach for cross-modal search in Section~\ref{sec:formulation}.
Section~\ref{sec:experiments} shows the experiments, followed by the concluding remarks in Section~\ref{sec:conclusion}.

\section{Related Work}
\label{sec:related work}
A variety of compact coding approaches for cross-modal similarity search have been developed over the last decade, which include hashing based approaches, quantization based approaches, and more recently deep learning based approaches. We briefly review some work related to our proposed method in this paper.

\subsection{Cross-modal Hashing}
Cross-modal hashing schemes~\cite{peng2013learning} aim at conducting fast similarity search across data from different modalities, which are also similar to the link prediction~\cite{qi2013link}. In these methods,
multi-modal data are usually embedded into a common Hamming space so that hash codes of different modalities can be directly compared
using the Hamming distance. Such an embedding can be considered as a hash function acting on input data trying to preserve
some underlying similarities. The main challenge in cross-modal hashing lies in how to exploit and build the intrinsic
relationships between multiple modalities.

Recently, various cross-modal hashing methods have been proposed. Sensitive hashing (CMSSH)~\cite{bronstein2010data} aligns similarities between points in a Hamming space shared across different modalities,
where the similarity between a pair of embedded data points is expressed as a superposition of weak classifier. Semantics
preserving hashing (SePH)~\cite{lin2015semantics} transforms the provided semantic affinities to a probability distribution and
approximates it with hash codes in a Hamming space. Considering the learned hash codes as supervised information, SePH learns the hash functions by kernel logistic regression. Co-regularized hashing (CRH)~\cite{zhen2012co} learns hash functions by
minimizing an intra-modality loss and an inter-modality loss. Inter-media hashing (IMH)~\cite{song2013inter} and cross-view
hashing (CVH)~\cite{kumar2011learning} extend spectral hashing to multi-modal setting. Meanwhile, some endeavors, such as
multimodal similarity-preserving hashing~\cite{masci2014multimodal}, sparse hashing~\cite{wu2014sparse}, composite hashing~\cite{zhang2011composite}, and collective matrix factorization~\cite{ding2014collective}, have been made to exploit similarity
relationships to learn hash codes accounting for different modalities.

\subsection{Cross-modal Quantization}
Quantization based methods try to approximate original data using some quantizers. Recently, many kinds of quantization techniques have been proposed including product quantization~\cite{jegou2011product}, additive quantization~\cite{babenko2014additive} and composite quantization~\cite{zhang2014composite}. Product quantization splits the input vector into distinct subvectors which are quantized
separately using distinct quantizers. Additive quantization approximates the input vectors using sums of several codewords from
different codebooks. Composite quantization is similar to additive quantization and introduces some extra constraints on the inner product between
codewords from different codebooks to further improve the query efficiency.

Currently, based on these quantization techniques, some approaches~\cite{wu2015quantized,long2016composite,zhang2016collaborative} were proposed for cross-modal
similarity search. Quantized correlation hashing~\cite{wu2015quantized} simultaneously learns hash functions and quantization
of hash codes by minimizing the inter-modality similarity disagreement as well as the binary quantization over each modality. Compositional
correlation quantization (CCQ)~\cite{long2016composite} transforms different modalities to an
isomorphic latent space, and then quantizes the isomorphic latent features into compact binary
codes by learning compositional quantizers. Collaborative quantization~\cite{zhang2016collaborative} jointly learns quantizers for
both modalities through aligning the quantized representations for each pair of image and text belonging to a document.
Besides, to make the representations in different modalities comparable, collaborative quantization simultaneously learns the common space for both modalities via matrix factorization,
and then conducts quantization to enable efficient and effective search via the Euclidean distance in the common space.

\subsection{Deep Learning Based Cross-modal Methods}
Deep learning has revolutionized computer vision~\cite{qi2017loss}, machine learning and other related areas, and has also demonstrated its effectiveness in improving accuracy of cross-modal hashing~\cite{jiang2016deep,cao2016deep} and cross-modal quantization~\cite{cao2017collective}. Deep cross-model hashing
(DCMH)~\cite{jiang2016deep} exploits pair-wise labels across different modalities to encourage semantic similar data points to have
similar hash codes and semantic dissimilar data points to have dissimilar hash codes. Moreover, DCMH integrates feature learning and
hash code learning into an unified deep framework. Deep visual-semantic hashing (DVSH)~\cite{cao2016deep} utilizes
a convolution neural network for image modality and a recurrent neural network for text modality. It first generates unified compact hash codes by combining information
from images and sentences, and then tries to approximate these learned hash codes by training a CNN for image modality and a RNN for text modality. After training, hash codes for
instances from different modalities can be directly generated from the corresponding network. Collective deep quantization (CDQ)~\cite{cao2017collective} solves the compact codes learning problem by using
a Bayesian learning framework, and learns deep representations and quantizers for all modalities by jointly optimizing these two parts.

Unlike these existing deep cross-modal methods~\cite{cao2017collective,cao2016deep,jiang2016deep}, SPDQ is the first attempt to
explicitly and jointly model a private subspace for each modality and a shared subspace between different modalities. Additionally, in the process of quantization, we utilize label alignment to greatly improve the quantization quality. Finally, we incorporate these two parts into an end-to-end architecture.

%------------------------------------------------------------------------
\section{Formulation}
\label{sec:formulation}

Suppose we have a database $\mathcal{D}$ with two modalities, $\mathcal{D} = \{{(\mathbf{X}_{i},\mathbf{X}_{t}),\mathbf{L}}\}$,
where $\mathbf{X}_{i} = \{\mathbf{x}_{i1},\mathbf{x}_{i2},...,\mathbf{x}_{iN}\} $ are data points from image modality,
$\mathbf{X}_{t}=\{\mathbf{x}_{t1},\mathbf{x}_{t2},...,\mathbf{x}_{tN}\}$ are data points from text modality, $\mathbf{L}=\{\mathbf{l}_{1},\mathbf{l}_{2},...,\mathbf{l}_{N}\}$ are their corresponding labels with $\mathbf{l}_{n}\in {\{0,1\}}^{K}$, $N$ is the number of data points, and $K$ is the number of classes. $\mathbf{X}_{i}$ can also be formulated as $\mathbf{X}_{i} = \{\mathbf{X}_{i}^{1},\mathbf{X}_{i}^{2},...,\mathbf{X}_{i}^{K}\} $, where $\mathbf{X}_{i}^{k} = \{\mathbf{x}^k_{i1},\mathbf{x}^k_{i2},...,\mathbf{x}^k_{iN_k}\}$ are image data points belonging to the $k$-th class, ${N}_{k}$ is the data point number belonging to the $k$-th class. $\mathbf{X}_{t}^{k}$ for text modality is defined similarly.

\begin{figure*}[!ht]
\centering
\includegraphics[width=0.97\textwidth]{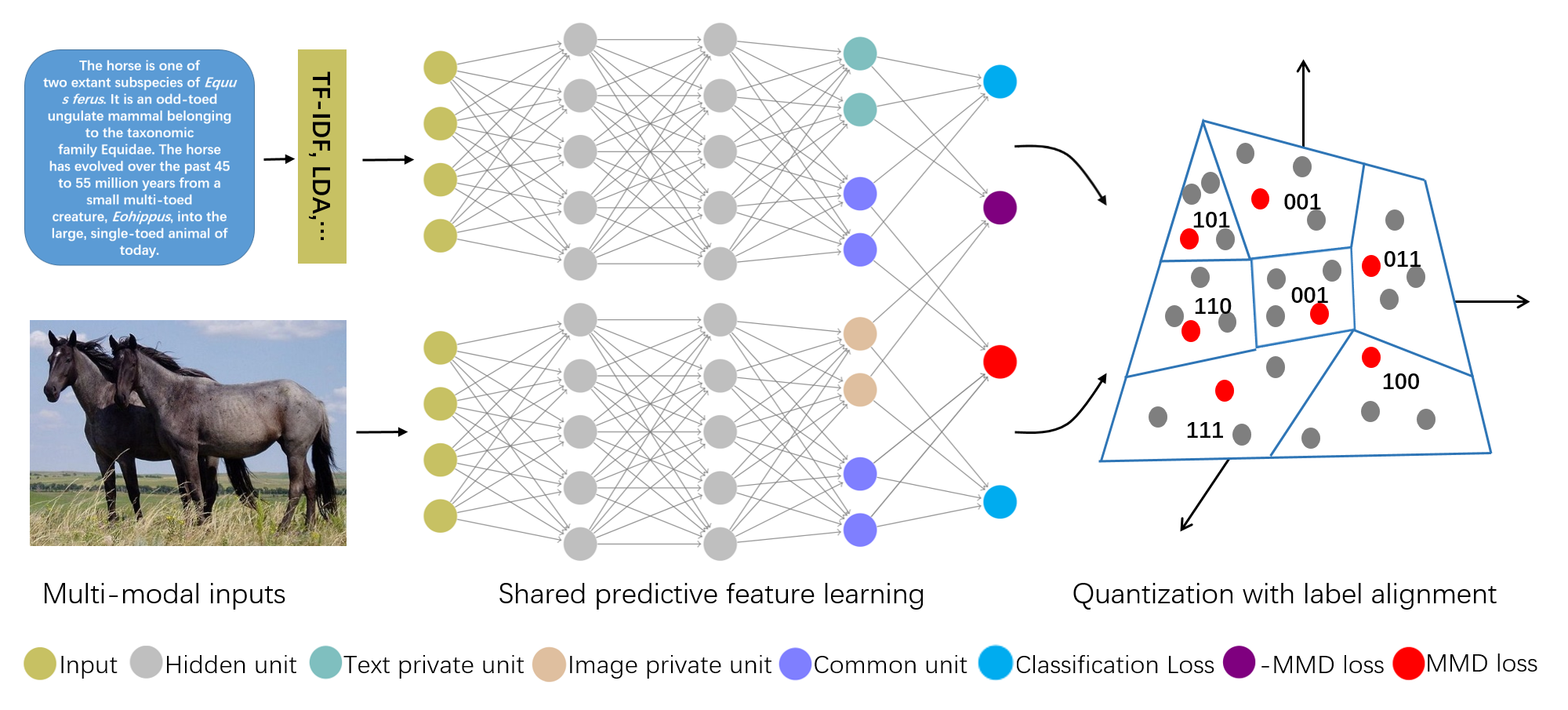}
\caption{Framework of the proposed method, where text private units and image private units exist in their corresponding modality-specific subspaces and common units exist in the shared subspace.}
\label{fig: model}
\end{figure*}

 Given an image (text) query $\mathbf{x}_{iq}$ $\left(\mathbf{x}_{tq} \right)$, the goal of cross-modal similarity search is to retrieve the closest matches in the text (image) database. In this paper, we first exploit convolutional neural networks (CNNs) to learn a shared subspace across different modalities and a private subspace for each modalities. Subsequently, in the shared subspace, we learn common representations by using label alignment and adopt additive quantization~\cite{babenko2014additive} to obtain the compact codes. The resulting deep network model is depicted in Fig.~\ref{fig: model}.

\subsection{Shared Predictive Representation Learning}
In cross-modal similarity search, representations of two related data points (e.g., an image and an associated text)
from different modalities should contain  some shared components since they describe the same object. Meanwhile, they should also
have private components since they come from different modalities. To achieve accurate search results, the intuitive
way is to simply preserve shared components among these representations as much as possible and ignore their private components.
However, private components are coupled with shared components in the original representation. According to the idea of subspace
decomposition~\cite{bousmalis2016domain}~\cite{deng2015multi} and component analysis~\cite{lu2016tensor}, explicitly modeling private components can enhance the ability to capture
shared components across different modality, thus can achieve more accurate searching. So, in this paper we explicitly model shared components and private components simultaneously.

Concretely, we project the multi-modal data points into
a shared subspace and two private subspaces, where the shared components and the private components are captured separately. And we utilize multiple kernel maximum
mean discrepancy (MK-MMD)~\cite{gretton2012optimal} as the distance metric in those subspaces. MK-MMD is a kernel based distance function first proposed for two sample test,
and has been successfully used in transfer learning~\cite{baktashmotlagh2016distribution} and domain adaptation~\cite{long2015learning}. In the
following, we first give a brief introduction to MK-MMD distance and then elaborate the learning process of the shared subspace and the private subspace.

Let $x$ and $y$ be random variables from distribution $p$ and $q$. The empirical estimate of the distance between $p$ and $q$, as defined by MK-MMD, is

\begin{equation}
\begin{aligned}
{d}_{k}^{2}(p,q)={\left \| \mathbb{E}_{x\sim p}[\phi ({x})]-\mathbb{E}_{y\sim q}[\phi ({y})]\right \|}_{{\mathcal{H}}_{k}}^{2},
\end{aligned}
\label{eqn: eq1}
\end{equation}
where ${\mathcal{H}}_{k}$ is a universal reproducing kernel Hilbert space (RKHS) endowed with a characteristic kernel $k$, $\phi$ is the mapping from original
data points to ${\mathcal{H}}_{k}$, and we have $k(x,y) = \left \langle \phi (x),\phi (y) \right \rangle$. For MK-MMD, the characteristic kernel $k$ is a convex combination of $m$ positive semi-definite kernels ${{k}_{u}}$

\begin{equation}
\begin{aligned}
\mathcal{K} :=  \left\{ k=\mathop{\sum}_{u=1}^{m}{\beta}_{u}{k}_{u}:\mathop{\sum}_{u=1}^{m}{\beta}_{u}=1,{\beta}_{u}\geq 0,\forall u \right\},
\end{aligned}
\label{eqn: eq2}
\end{equation}
where the coefficients $\beta_{u}$ are constrained to guarantee the derived multi-kernel $k$ characteristic. Having a range of kernels is beneficial since the
distributions of the features change during learning and different components of the multi-kernel might be responsible at different times, so that we can get a large
distance when the distributions are not similar.

In this paper, we exploit MK-MMD as the distance function for learning representations in the shared subspace and the private subspace.
As analyzed before, the shared subspace captures common components between different modalities, and the private subspace captures unique components for each modality. So in the shared subspace, representations of data points with the same class from different modalities should be as
similar as possible, and in the private subspace, the representations from different modalities reflecting modality-specific information should be as dissimilar as possible. Thus, we can get
the following loss function
\begin{equation}
\begin{aligned}
\mathcal{L}_{1} %&= %{\alpha}_{c}{d}_{k}^{2}({p}_{{c}_{1}},{p}_{{c}_{2}}) - {\alpha}_{s}{d}_{k}^{2}({p}_{{s}_{1}},{p}_{{s}_{2}})\\
        &= \mathop{\sum}_{k=1}^{K}\{{\left \| \mathbb{E}[\phi (\mathbf{s}_{i}^{k})]-\mathbb{E}[\phi (\mathbf{s}_{t}^{k})]\right \|}_{{\mathcal{H}}_{k}}^{2}\\
        &- {\left \| \mathbb{E}[\phi (\mathbf{r}_{i}^{k})]-\mathbb{E}[\phi (\mathbf{r}_{t}^{k})]\right \|}_{{\mathcal{H}}_{k}}^{2}\},
\end{aligned}
\label{eqn: eq3}
\end{equation}
where $\mathbf{s}_{i}^{k}$ and $\mathbf{s}_{t}^{k}$ are representations for $\mathbf{x}_{i}^{k}$ and $\mathbf{x}_{t}^{k}$ in the shared subspace, $\mathbf{r}_{i}^{k}$ and $\mathbf{r}_{t}^{k}$ are the corresponding representations in their private subspaces.

To ensure that the modality-specific representations are useful (avoiding trivial solutions) and enhance the discriminative
ability of representations in both private subspace and shared subspace, for each modality we concatenate the representations learned from shared subspace and private subspace into a complete
representation as $\mathbf{h}_{i} = [\mathbf{s}_{i},\mathbf{r}_{i}]$ and $\mathbf{h}_{t} = [\mathbf{s}_{t},\mathbf{r}_{t}]$.
Then, we introduce a classification loss for these complete representations as

\begin{equation}
\begin{aligned}
\mathcal{L}_{2} = \mathcal{L}_{c}(\mathbf{h}_{i},\mathbf{l})+\mathcal{L}_{c}(\mathbf{h}_{t},\mathbf{l}),
%{L}_{2} &= -\frac{1}{n}\sum_{n=1}^{N}[{y}_{n}log(\frac{1}{1+exp(-{x}^{n})})\\
%& + (1-{y}_{n})log(\frac{exp(-{x}^{n})}{1+exp(-{x}^{n})})]
\end{aligned}
\label{eqn: eq4}
\end{equation}
where $\mathcal{L}_{c}$ is the sigmoid cross-entropy loss, $\mathbf{l}$ is the corresponding label vector.

Combine~\eqref{eqn: eq3} and~\eqref{eqn: eq4}, we can formulate the overall objective function for shared predictive representation learning as

\begin{equation}
\begin{aligned}
\mathcal{O}_{l} = \mathcal{L}_{1}+\alpha\mathcal{L}_{2},
%{L}_{2} &= -\frac{1}{n}\sum_{n=1}^{N}[{y}_{n}log(\frac{1}{1+exp(-{x}^{n})})\\
%& + (1-{y}_{n})log(\frac{exp(-{x}^{n})}{1+exp(-{x}^{n})})]
 \end{aligned}
 \label{eqn: eq5}
 \end{equation}
where $\alpha$ is the weight to balance these two parts.

For each modality, we design a neural network to learn the shared and private representations as depicted in Fig.~\ref{fig: model}, which will be elaborated in Section~\ref{sec:experiments}. By utilizing the designed networks and optimizing the objective function $\mathcal{O}_{l}$, we can explicitly map data points from different modalities into a shared subspace and their modality-specific private subspaces. The networks for the shared subspace learning can be used to predict shared representations for data points from different modalities. Since shared representations contain all the correlated information across different modalities which are crucial for cross-modal retrieval and private components contain only modality-specific information which are inapplicable to cross-modal retrieval, so for the following quantization procedure, we only
use the representations in the shared subspace.

\subsection{Quantization with Label Alignment}
If multi-modal data points sharing the same semantic meaning (belong to same categories), it is natural to assume that they share some common structures which are correlated with their labels. Therefore, to preserve the semantic similarity and reduce the intra-class diversity, we assume that representations in shared subspace are composed of category-specific structures which reflect the shared characteristics inside heterogeneous data points, and we use label information to discover and preserve these category-specific structures. The objective function can be defined as

\begin{equation}
\begin{aligned}
{\mathcal{L}}_{3}=&{\left \| \mathbf{C}_{i}\mathbf{S}_{i}-\mathbf{Z}\mathbf{L} \right \|}^{2}_{F}+{\left \| \mathbf{C}_{t}\mathbf{S}_{t}-\mathbf{ZL} \right \|}^{2}_{F}\\
&s.t.\; \mathbf{C}_{i}\mathbf{C}_{i}^{T} = \mathbf{I},\mathbf{C}_{t}\mathbf{C}_{t}^{T} = \mathbf{I},
 \end{aligned}
\label{eqn: eq6}
\end{equation}
where $\mathbf{S}_{i} = [\mathbf{s}_{i1},\mathbf{s}_{i1},...,\mathbf{s}_{iN}]$ are the representations for image data points in the shared subspace, $\mathbf{S}_{t} = [\mathbf{s}_{t1},\mathbf{s}_{t1},...,\mathbf{s}_{tN}]$ are the representations for text data points in the shared subspace, and $\mathbf{C}_{i}$ and $\mathbf{C}_{t}$ are the transformation matrixes that align $\mathbf{S}_i$ and $\mathbf{S}_t$ in the label space.
In addition, the constraints $\mathbf{C}_{i}\mathbf{C}_{i}^{T} = \mathbf{I}, \mathbf{C}_{t}\mathbf{C}_{t}^{T} = \mathbf{I}$ are used to make $\mathbf{C}_i$ and $\mathbf{C}_t$ be orthogonal projections. We construct an auxiliary matrix $\mathbf{Z}=\left [ \mathbf{z}_{1}, \mathbf{z}_{2},...\mathbf{z}_{K} \right ]$ to reflect the category-specific structures,  where $\mathbf{z}_i$ is the representation for $i$-th class in the label space. Since all data points sharing the same labels have the same $\mathbf{ZL}$, by optimizing ${\mathcal{L}}_{3}$ we can explicitly reduce the intra-class diversity, and $\mathbf{ZL}$ can be regarded as the surrogate of $\mathbf{C}_{i}\mathbf{S}_i$ and $\mathbf{C}_{t}\mathbf{S}_t$ in the label space.

Based on the surrogate representations $\mathbf{ZL}$, we adopt additive quantization~\cite{babenko2014additive} to learn the final compact codes. Additive quantization aims to approximate representations as the sum of $M$ elements selected from $M$ dictionaries. We can approximate $\mathbf{ZL}$ by minimizing

\begin{equation}
\begin{aligned}
{\mathcal{L}}_{4}={\|\mathbf{ZL} - \mathbf{DB}\|}_{F}^{2}.
\end{aligned}
\label{eqn: eq7}
\end{equation}
Here $\mathbf{D} = [\mathbf{D}_{1},\mathbf{D}_{2},...\mathbf{D}_{M}]$ corresponds to $M$ dictionaries, $\mathbf{D}_{m} = [\mathbf{d}_{m1},\mathbf{d}_{m2},...,\mathbf{d}_{mK}]$ corresponds to the $m$-th dictionary  with $K$ elements, and $\mathbf{B} = [\mathbf{B}_{1},\mathbf{B}_{2},...,\mathbf{B}_{N}]$ with $\mathbf{B}_{n} = [\mathbf{b}_{n1},\mathbf{b}_{n2},...,\mathbf{b}_{nM}]$ is the indicator matrix, where each $\mathbf{b}_{nm}$ is a 1-of-$K$ binary vector indicating which one (and only one) of the $K$ dictionary elements is selected to approximate the data points.

Combine~\eqref{eqn: eq6} and~\eqref{eqn: eq7}, we obtain the overall objective function for quantization as
\begin{equation}
\begin{aligned}
\mathcal{O}_{q}&=\mathcal{L}_{3} + \beta\mathcal{L}_{4}\\
&= {\left \| \mathbf{C}_{i}\mathbf{S}_{i}-\mathbf{ZL} \right \|}^{2}_{F}+{\left \| \mathbf{C}_{t}\mathbf{S}_{t}-\mathbf{ZL} \right \|}^{2}_{F}\\
& +\beta{\left \|\mathbf{ZL}-\mathbf{DB} \right \|}^{2}_{F}\\
& s.t.\; \mathbf{C}_{i}\mathbf{C}_{i}^{T} = \mathbf{I},\mathbf{C}_{t}\mathbf{C}_{t}^{T} = \mathbf{I},
\end{aligned}
\label{eqn: eq8}
\end{equation}
where $\beta$ are weights to balance the two parts.

\subsection{Overall Objective Function}
When merging~\eqref{eqn: eq5} and~\eqref{eqn: eq8} together, the overall objective function for learning shared predictive representations and quantizing these representations into compact codes with label alignment can be expressed as

\begin{equation}
\begin{aligned}
\mathcal{O}= \mathcal{O}_{l} + \lambda\mathcal{O}_{q}\\
%&= \lambda(\mathop{\sum}_{k=1}^{K}\{{\left \| {E}[\phi ({S}_{i}^{k})]-{E}[\phi ({S}_{t}^{k})]\right \|}_{{\mathcal{H}}_{k}}^{2}\\
%&+ {\alpha}{\left \| {E}[\phi ({P}_{i}^{k})]-{E}[\phi ({P}_{t}^{k})]\right \|}_{{\mathcal{H}}_{k}}^{2}\}\\
%&+ \beta(\mathcal{L}_{c}({H}_{i},L)+\mathcal{L}_{c}({H}_{t},L)))\\
%&+ (1-\lambda)({\left \| {C}_{i}{S}_{i}-ZL \right \|}^{2}_{F}+{\left \| {C}_{t}{S}_{t}-ZL \right \|}^{2}_{F}\\
%& +\gamma{\left \|ZL-DB \right \|}^{2}_{F})\\
 s.t.\; \mathbf{C}_{i}\mathbf{C}_{i}^{T} = \mathbf{I}, \mathbf{C}_{t}\mathbf{C}_{t}^{T} = \mathbf{I},
\end{aligned}
\label{eqn: eq9}
\end{equation}
where $\lambda$ is the weight to balance the representation learning and quantization parts. By minimizing~\eqref{eqn: eq9}, we can jointly
optimize these two parts which will help to learn more suitable representations for quantization and generate more effectively compact codes.

\subsection{Search Process}
Approximate nearest neighbor (ANN) search based on inner product distance is a powerful technique for quantization methods. Given an image query $\mathbf{x}_{iq}$, we first compute its common representation $\mathbf{s}_{iq}$ using the trained image modality network, and obtain its corresponding representation in the label space denoted as $\mathbf{C}_{i}\mathbf{s}_{iq}$. Similarly to~\cite{cao2017collective}, we use Asymmetric Quantizer Distance (AQD) to calculate the distance between the query $\mathbf{x}_{iq}$ and text point $\mathbf{x}_{td}$ in database as

\begin{equation}
\begin{aligned}
AQD(\mathbf{x}_{iq}, \mathbf{x}_{td}) = {\mathbf{C}_{i}\mathbf{s}_{iq}}^{\top}\cdot(\mathop{\sum}_{m=1}^{M} \mathbf{D}_{m}\mathbf{b}_{dm})
\end{aligned}
\label{eqn: eq10}
\end{equation}
where $\mathbf{DB}_d = \sum^M_{m = 1} \mathbf{D}_{m}\mathbf{b}_{dm}$ is the quantized representation for text database point $\mathbf{x}_{ti}$. Given a query, the inner products for all $M$ dictionaries and all $K$ possible values of $\mathbf{b}_{im}$ can be pre-computed and stored in a $M \times K$ lookup table, which is used to compute AQD between the query and all database points. Each
AQD calculation entails $M$ table lookups and $M$ additions, and is slightly more costly than Hamming distance.

\section{Optimization}
\label{sec:optimization}
We optimize the proposed problem by alternatively solving two sub-problems: updating network parameters with quantization parameters fixed, and updating quantization parameters with the network parameters fixed.

1) Fix quantization parameters $\mathbf{C}_{i}$, $\mathbf{C}_{t}$, $\mathbf{Z}$, $\mathbf{D}$ and $\mathbf{B}$, update network parameters. In this paper, we adopt the unbiased estimate of MK-MMD~\cite{gretton2012optimal}.
Specifically,

\begin{equation}
\begin{aligned}
d_k^2\left( p, q \right) = \frac{2}{N}\mathop{\sum}_{n=1}^{{N}/{2}}\eta\left( \mathbf{u}_n \right),
\end{aligned}
 \label{eqn: eq11}
\end{equation}
where, for shared representations, quad-tuple $\mathbf{u}_{n} \triangleq(\mathbf{s}_{i(2n-1)},\mathbf{s}_{i(2n)},\mathbf{s}_{t(2n-1)},\mathbf{s}_{t(2n-1)})$, $N$ is the number of points in a mini-batch, and
\begin{equation}
\begin{aligned}
%\begin{split}
{\eta}(\mathbf{u}_{n}) &= k(\mathbf{s}_{i(2n-1)},\mathbf{s}_{i(2n)}) - k(\mathbf{s}_{i(2n-1)},\mathbf{s}_{t(2n)})\\
 & + k(\mathbf{s}_{t(2n-1)}, \mathbf{s}_{t(2n)})- k(\mathbf{s}_{t(2n-1)},\mathbf{s}_{i(2n)}).
%\end{split}
\end{aligned}
\label{eqn: eq12}
\end{equation}

The distance between private representations is similar to the shared representations. When we optimize the network parameters by mini-batch SGD, we only need to consider the gradients of~\eqref{eqn: eq9} with respect to each data point. Actually we only need to compute the gradients of ${\eta}(\mathbf{u}_{n})$ for the quad-tuple $\mathbf{u}_{n}=(\mathbf{s}_{i(2n-1)},\mathbf{s}_{i(2n)},\mathbf{s}_{t(2n-1)},\mathbf{s}_{t(2n-1)})$. Given kernel $k$ as the convex combination of $m$ Gaussian kernels $\{k_{a}(\mathbf{x}_{i}, \mathbf{x}_{j}) = {e}^{-{\left \|\mathbf{x}_{i} - \mathbf{x}_{j} \right\|}^{2}/{\tau}_{a}}\}$, since:

\begin{equation}
\begin{aligned}
\frac{\partial{k(\mathbf{s}_{i(2n-1)},\mathbf{s}_{i(2n)})}}{\partial \mathbf{s}_{i(2n)}} = &-{\sum}_{a=1}^{m}\frac{2{\beta}_{a}}{{\tau}_{a}}{k}_{a}(\mathbf{s}_{i(2n-1)},\mathbf{s}_{i(2n)})\\
&\times (\mathbf{s}_{i(2n-1)} - \mathbf{s}_{i(2n)})
\end{aligned}
\label{eqn: eq13}
\end{equation}

Combine~\eqref{eqn: eq1} and~\eqref{eqn: eq3}, the gradient $\frac{\partial{\mathcal{L}_{1}}}{\partial\mathbf{s}_{i\cdot}}$ and $\frac{\partial{\mathcal{L}_{1}}}{\partial\mathbf{r}_{i\cdot}}$ can be readily computed.
Since we adopt sigmoid cross-entropy loss for the classification loss in~\eqref{eqn: eq4}, the gradient $\frac{\partial{\mathcal{L}_{2}}}{\partial\mathbf{s}_{in}}$ and $\frac{\partial{\mathcal{L}_{2}}}{\partial\mathbf{r}_{in}}$
can also be readily computed. Then we can get

\begin{equation}
\begin{aligned}
\frac{\partial{\mathcal{O}}}{\partial\mathbf{r}_{in}} = \frac{\partial\mathcal{L}_{1}}{\partial\mathbf{r}_{in}} + \alpha\frac{\partial{\mathcal{L}_{2}}}{\partial\mathbf{r}_{in}},\\
\frac{\partial{\mathcal{O}}}{\partial\mathbf{r}_{tn}} = \frac{\partial\mathcal{L}_{1}}{\partial\mathbf{r}_{tn}} + \alpha\frac{\partial{\mathcal{L}_{2}}}{\partial\mathbf{r}_{tn}}.
\end{aligned}
\label{eqn: eq14}
\end{equation}
Fix $\mathbf{C}_{i}$, $\mathbf{C}_{t}$ and $\mathbf{Z}$, we can get
\begin{equation}
\begin{aligned}
 \frac{\partial{\mathcal{O}_{q}}}{\partial\mathbf{s}_{in}} = 2 \mathbf{C_i}^{T}(\mathbf{C}_{i}\mathbf{s}_{in}-\mathbf{Zl}_n),\\
 \frac{\partial{\mathcal{O}_{q}}}{\partial\mathbf{s}_{tn}} = 2 \mathbf{C_t}^{T}(\mathbf{C}_{t}\mathbf{s}_{tn}-\mathbf{Zl}_n).\\
\end{aligned}
\label{eqn: eq15}
\end{equation}
Thus we can compute the gradient of the overall objective function $\mathcal{O}$ with regard to $\mathbf{s}_{i}$ and $\mathbf{s}_{t}$ as
\begin{equation}
\begin{aligned}
 \frac{\partial{\mathcal{O}}}{\partial\mathbf{s}_{in}} = (\frac{\partial\mathcal{L}_{1}}{\partial\mathbf{s}_{in}} + \alpha\frac{\partial{\mathcal{L}_{2}}}{\partial\mathbf{s}_{in}}) + \lambda(\frac{\partial{\mathcal{O}_{q}}}{\partial\mathbf{s}_{in}}),\\
 \frac{\partial{\mathcal{O}}}{\partial\mathbf{s}_{tn}} = (\frac{\partial\mathcal{L}_{1}}{\partial\mathbf{s}_{tn}} + \alpha\frac{\partial{\mathcal{L}_{2}}}{\partial\mathbf{s}_{tn}}) + \lambda(\frac{\partial{\mathcal{O}_{q}}}{\partial\mathbf{s}_{tn}}).
\end{aligned}
 \label{eqn: eq16}
 \end{equation}
 Then using~\eqref{eqn: eq14} and~\eqref{eqn: eq16}, we can update the parameters of image modality network and text modality network with chain rule.

2) Fix parameters of image modality network and text modality network, $\mathbf{C}_i$, $\mathbf{Z}$, when $\mathbf{S}_i$ are known, we update the transform matrix $\mathbf{C}_{i}$ by
 \begin{equation}
\begin{aligned}
&{\min}_{\mathbf{C}_{i}}{\left \| \mathbf{C}_{i}\mathbf{S}_{i}-\mathbf{ZL} \right \|}_{F}^{2},\\
& s.t.\; \mathbf{C}_{i}\mathbf{C}_{i}^{T} = \mathbf{I}.
 \end{aligned}
 \label{eqn: eq17}
 \end{equation}
\eqref{eqn: eq17} can be considered as the Orthogonal Procrustes problem and is solved by SVD algorithms.
Specifically, we perform SVD as $\mathbf{(ZL)}\mathbf{{S}_{i}}^{T} = \mathbf{U}_{i}\mathbf{Y}_{i}\mathbf{W}^{T}_{i}$, and then we can update $\mathbf{C}_{i}$ by:
 \begin{equation}
\begin{aligned}
\mathbf{C}_{i}=\mathbf{U}_{i}\mathbf{W}^{T}_{i}.
 \end{aligned}
 \label{eqn: eq18}
 \end{equation}

3) Fix parameters of image modality network and text modality network, $\mathbf{C}_t$ and $\mathbf{Z}$. Similarly to $\mathbf{C}_i$, Given $\mathbf{(ZL)}\mathbf{{S}_{t}}^{T} = \mathbf{U}_{t}\mathbf{Y}_{t}\mathbf{W}^{T}_{t}$, We can update the transform matrix $\mathbf{C}_{t}$ by
 \begin{equation}
\begin{aligned}
\mathbf{C}_{t}= \mathbf{U}_{t}\mathbf{W}^{T}_{t}.
 \end{aligned}
 \label{eqn: eq19}
 \end{equation}

 4) Fix $\mathbf{C}_i$, $\mathbf{C}_t$, $\mathbf{D}$ and $\mathbf{B}$, let the derivation of $O$ with regard to $\mathbf{Z}$ equals to zero, we can get
 \begin{equation}
\begin{aligned}
\mathbf{Z} = {[(\mathbf{C}_{i}\mathbf{S}_{i} + \mathbf{C}_{t}\mathbf{S}_{t}) +\beta \mathbf{DB}]\mathbf{L}}^{T}{((2+\beta)(\mathbf{L}\mathbf{L}^{T}))}^{-1},
\end{aligned}
 \label{eqn: eq20}
\end{equation}

5) Fix $\mathbf{Z}$ and $\mathbf{B}$, and neglect the irrelevant items, we can update the dictionary $\mathbf{D}$ by optimizing
 \begin{equation}
\begin{aligned}
&{\min}_\mathbf{D}{\left \| \mathbf{DB-ZL} \right \|}_{F}^{2}.
 \end{aligned}
 \label{eqn: eq21}
 \end{equation}
Eq.~\eqref{eqn: eq21} is an unconstrained quadratic problem with analytic solution, so we can update $\mathbf{D}$ by
 \begin{equation}
\begin{aligned}
D=[\mathbf{ZL}\mathbf{B}^{T}]{[\mathbf{B}\mathbf{B}^{T}]}^{-1}.
 \end{aligned}
 \label{eqn: eq22}
 \end{equation}

6) Fix $\mathbf{D}$ and $\mathbf{Z}$, since each $\mathbf{b}_{n}$ is independent on ${\mathbf{b}_{{n}^{'}}}\left( {{n}^{'}}\not=n\right) $, the optimization problem for $B$ is decomposed to $N$ subproblems,
\begin{equation}
\begin{aligned}
{\min}_{\mathbf{b}_{n}}{\left \|\mathbf{Z}\mathbf{l}_{n}-{\sum}_{m=1}^{M}\mathbf{D}_{m}\mathbf{b}_{nm}\right \|}^{2}\\
s.t.{\left \|\mathbf{b}_{nm}\right \|}_{0} = 1, \mathbf{b}_{nm}\in {\{0,1\}}^{K}.
\end{aligned}
 \label{eqn: eq23}
\end{equation}
This optimization problem in~\eqref{eqn: eq23} is generally NP-hard.
We approximately solve this problem by the Iterated Conditional Modes (ICM) algorithm.
Fixing ${\{\mathbf{b}_{n{m}^{'}}}\}_{{m}^{'}\not= m}$, $\mathbf{b}_{nm}$ is updated by exhaustively
checking all the elements in $\mathbf{D}_{m}$, finding the element such that the objective function
is minimized, and setting the corresponding entry of $\mathbf{b}_{nm}$ to be 1 and all the others
to be 0. This algorithm is guaranteed to converge and can be terminated if maximum iterations are reached.

Altogether, SPDQ is summarized in Algorithm~\ref{alg: alg1}.

\begin{algorithm}[!t]
 \SetAlgoNoLine
 \caption{\small Shared predictive Deep quantization}
 \textbf{Training Stage}\\
 \BlankLine
 \textbf{Input:}{ Image $\mathbf{X}_i$ text $\mathbf{X}_t$, semantic labels $\mathbf{L}$, code length $k$, parameters $\alpha$, $\beta$, and $\lambda$.}\\
 \textbf{Output:}{ Parameters for image network and text network, parameters for quantization learning $\mathbf{C}_i$, $\mathbf{C}_t$, $\mathbf{Z}$, $\mathbf{D}$ and compact codes $\mathbf{B}$.}\\
 \textbf{Procedure:}\\
  1. Initialize parameters for image network and text network, initialize $\mathbf{C}_i$, $\mathbf{C}_t$, $\mathbf{Z}$, $\mathbf{D}$ and $\mathbf{B}$ by random matrices, Mini-batch size
  $N_{i} = N_{t} = 128$.\\
  \Repeat{Until convergency}{
    2.1 Randomly sample $N_i$ points for $\mathbf{X}_i$ and $N_t$ points from $\mathbf{X}_t$ to construct a mini-batch.\\
    2.2 Calculate the outputs $\mathbf{s}_i$, $\mathbf{s}_t$, $\mathbf{r}_i$ and $\mathbf{r}_t$.\\
    2.3 Update parameters of image network by~\eqref{eqn: eq14} and~\eqref{eqn: eq16}.\\
    2.4 Update parameters of text network by~\eqref{eqn: eq14} and~\eqref{eqn: eq16}.\\
    2.5 Update $\mathbf{C}_i$ by~\eqref{eqn: eq18}.\\
    2.6 Update $\mathbf{C}_t$ by~\eqref{eqn: eq19}.\\
    2.7 Update $\mathbf{Z}$ by~\eqref{eqn: eq20}.\\
    2.8 Update $\mathbf{D}$ by~\eqref{eqn: eq22}.\\
    2.9 Update $\mathbf{B}$ by the ICM algorithm.}
 \BlankLine
 \textbf{Testing Stage}\\
 \BlankLine
 \textbf{Input:}{ Image query $\mathbf{x}_i$ or text query $\mathbf{x}_t$, parameters for image network and text network, $\mathbf{C}_i$, ${C}_t$, $\mathbf{D}$ and $\mathbf{B}$.}\\
 \textbf{Output:}{ Ranked neighbor list for query}\\
 \textbf{Procedure:}\\
 1. Calculate the shared representation $\mathbf{s}_i$ or $\mathbf{s}_t$ by forward the image network or text network.\\
 2. Calculate the distance between query point and the database by~\eqref{eqn: eq10}.
 \label{alg: alg1}
\end{algorithm}

%-------------------------------------------------------------------------
\section{Experiments}
\label{sec:experiments}

\begin{table*}[!t]
\newcommand{\tabincell}[2]{\begin{tabular}{@{}#1@{}}#2\end{tabular}}
\centering
\caption{Comparison with baselines with hand-crafted features in terms of MAP. The best accuracy is shown in boldface.}
\begin{tabular}{|c||c||c|c|c|c||c|c|c|c|}
\hline
\multirow{2}{*}{Task} &\multirow{2}{*}{method} &\multicolumn{4}{c||}{FLICKR25K}&\multicolumn{4}{c|}{NUS-WIDE}\\
\cline{3-10}
&  & 16 bits & 32 bits & 64 bits & 128 bits & 16 bits & 32 bits & 64 bits & 128 bits\\
\hline
\hline
\multirow{6}{*}{\tabincell{c}{Image Query \\ v.s.\\ Text Database}}
 %& CMFH  & 0.8445 & 0.8756 & 0.8852 & 0.9009 & 0.7454 & 0.7905 & 0.8128 & 0.8331 \\ \cline{2-10}
 & CMSSH & 0.6500 & 0.6052 & 0.6106 & 0.6667 & 0.7223 & 0.7354 & 0.7483 & 0.7532 \\ \cline{2-10}
 & SCM  & 0.6976 & 0.6875 & 0.7089 & 0.7186 & 0.6976 & 0.6854 & 0.7089 & 0.7186 \\ \cline{2-10}
 %& LSSH & 0.7685 & 0.8145 & 0.8538 & 0.8750 & 0.7179 & 0.7557 & 0.7873 & 0.8008 \\ \cline{2-10}
 %& STMH & 0.7355 & 0.7751 & 0.8315 & 0.8502 & 0.6403 & 0.7239 & 0.7824 & 0.7975\\ \cline{2-10}
 & CVH  & 0.5922 & 0.5900 & 0.5913 & 0.6369 & 0.6095 & 0.5917 & 0.5987 & 0.6211 \\ \cline{2-10}
 & SePH & 0.6214 & 0.6416 & 0.6433 & 0.6469 & 0.6225 & 0.6296 & 0.6377 & 0.6288 \\ \cline{2-10}
 & DCMH & 0.7576 & 0.7985 & 0.8152 & 0.8369 & 0.7353 & 0.7628 & 0.7805 & 0.7912  \\ \cline{2-10}
  & CDQ  & 0.9047 & 0.9094 & 0.9109 & 0.8587  & 0.7917 & 0.7978 & 0.8102 & 0.8214\\\cline{2-10}
 & Ours &\textbf{0.9443} &\textbf{0.9476} & \textbf{0.9482} & \textbf{0.8725}
 & \textbf{0.9276} & \textbf{0.9344} & \textbf{0.9251}& \textbf{0.9299} \\ \cline{2-10}
 \hline
 \hline
\multirow{6}{*}{\tabincell{c}{Text Query \\ v.s. \\ Image Database}}
  %& CMFH  & 0.8344 & 0.8495 & 0.8588 & 0.8676 & 0.7365 & 0.7626 & 0.7755 & 0.7864\\ \cline{2-10}
 & CMSSH & 0.5014 & 0.4988 & 0.5002 & 0.5015 & 0.6250 & 0.6315 & 0.6445 & 0.6616 \\ \cline{2-10}
 & SCM  & 0.5693 & 0.5710 & 0.5918 & 0.6028 & 0.5549 & 0.5914 & 0.5991 & 0.6007 \\ \cline{2-10}
 %& LSSH & 0.7122 & 0.7502 & 0.7655 & 0.7819 & 0.6307 & 0.6736 & 0.7023 & 0.7074 \\ \cline{2-10}
 %& STMH & 0.7149 & 0.7411 & 0.7443 & 0.7509 & 0.6442 & 0.6654 & 0.6819 & 0.6861 \\ \cline{2-10}
 & CVH  & 0.5352 & 0.5254 & 0.5011 & 0.4705 & 0.5601 & 0.5439 & 0.5160 & 0.4821 \\ \cline{2-10}
 & SePH & 0.5506 & 0.5686 & 0.5750 & 0.5837 & 0.5943 & 0.5835 & 0.6119 & 0.6320 \\ \cline{2-10}
 & DCMH & 0.7013 & 0.7288 & 0.7458& 0.7698 & 0.6898 & 0.7102 & 0.7358 & 0.7557  \\ \cline{2-10}
 & CDQ  & 0.8848 & 0.8768 & 0.8841 & 0.8736  & 0.8227 & 0.8184 & 0.8283 & 0.8201\\\cline{2-10}
  & Ours & \textbf{0.9278} &\textbf{0.9318} & \textbf{0.9313} & \textbf{0.9109}& \textbf{0.8914} & \textbf{0.8983} & \textbf{0.8959}& \textbf{0.8900} \\ \cline{2-10}
 \hline
\end{tabular}
\label{tab: tab11}
\end{table*}

\begin{table*}[!t]
\newcommand{\tabincell}[2]{\begin{tabular}{@{}#1@{}}#2\end{tabular}}
\centering
\caption{Comparison with baselines with CNN features in terms of MAP. The best accuracy is shown in boldface.}
\begin{tabular}{|c||c||c|c|c|c||c|c|c|c|}
\hline
\multirow{2}{*}{Task} &\multirow{2}{*}{method} &\multicolumn{4}{c||}{FLICKR25K}&\multicolumn{4}{c|}{NUS-WIDE}\\
\cline{3-10}
&  & 16 bits & 32 bits & 64 bits & 128 bits & 16 bits & 32 bits & 64 bits & 128 bits\\
\hline
\hline
\multirow{6}{*}{\tabincell{c}{Image Query \\ v.s.\\ Text Database}}
 %& CMFH  & 0.8445 & 0.8756 & 0.8852 & 0.9009 & 0.7454 & 0.7905 & 0.8128 & 0.8331 \\ \cline{2-10}
 & CMSSH & 0.6484 & 0.6484 & 0.6784 & 0.5967 & 0.7074 & 0.7346 & 0.7353 & 0.7450 \\ \cline{2-10}
 & SCM  & 0.7749 & 0.7868 & 0.8012 & 0.8024 & 0.7885 & 0.7970 & 0.8131 & 0.8209 \\ \cline{2-10}
 %& LSSH & 0.7685 & 0.8145 & 0.8538 & 0.8750 & 0.7179 & 0.7557 & 0.7873 & 0.8008 \\ \cline{2-10}
 %& STMH & 0.7355 & 0.7751 & 0.8315 & 0.8502 & 0.6403 & 0.7239 & 0.7824 & 0.7975\\ \cline{2-10}
 & CVH  & 0.7100 & 0.6776 & 0.6438 & 0.6154 & 0.7157 & 0.6988 & 0.6573 & 0.5993 \\ \cline{2-10}
 & SePH & 0.7991 & 0.8199 & 0.8435 & 0.8480 & 0.7893 & 0.8052 & 0.8198 & 0.8333 \\ \cline{2-10}
 & DCMH & 0.7576 & 0.7985 & 0.8152 & 0.8369 & 0.7353 & 0.7628 & 0.7805 & 0.7912  \\ \cline{2-10}
 & CDQ  & 0.9047 & 0.9094 & 0.9109 & 0.8587  & 0.7917 & 0.7978 & 0.8102 & 0.8214   \\\cline{2-10}
 & Ours &\textbf{0.9443} &\textbf{0.9476} & \textbf{0.9482} & \textbf{0.8725}
 & \textbf{0.9276} & \textbf{0.9344} & \textbf{0.9251}& \textbf{0.9299} \\ \cline{2-10}
 \hline
 \hline
\multirow{6}{*}{\tabincell{c}{Text Query \\ v.s. \\ Image Database}}
  %& CMFH  & 0.8344 & 0.8495 & 0.8588 & 0.8676 & 0.7365 & 0.7626 & 0.7755 & 0.7864\\ \cline{2-10}
 & CMSSH & 0.5890 & 0.5752 & 0.5795 & 0.5714 & 0.6556 & 0.6749 & 0.7065 & 0.7230 \\ \cline{2-10}
 & SCM  & 0.6624 & 0.6663 & 0.6837 & 0.6880 & 0.6922 & 0.6854 & 0.7140 & 0.7244 \\ \cline{2-10}
 %& LSSH & 0.7122 & 0.7502 & 0.7655 & 0.7819 & 0.6307 & 0.6736 & 0.7023 & 0.7074 \\ \cline{2-10}
 %& STMH & 0.7149 & 0.7411 & 0.7443 & 0.7509 & 0.6442 & 0.6654 & 0.6819 & 0.6861 \\ \cline{2-10}
 & CVH  & 0.7220 & 0.6522 & 0.5972 & 0.5784 & 0.7085 & 0.7383 & 0.7553 & 0.7475 \\ \cline{2-10}
 & SePH & 0.7454 & 0.7421 & 0.7614 & 0.7893 & 0.6947 & 0.7062 & 0.7235 & 0.7251 \\ \cline{2-10}
 & DCMH & 0.7013 & 0.7288 & 0.7458& 0.7698 & 0.6898 & 0.7102 & 0.7358 & 0.7557  \\ \cline{2-10}
 & CDQ  & 0.8848 & 0.8768 & 0.8841 & 0.8736  & 0.8227 & 0.8184 & 0.8283 & 0.8201   \\\cline{2-10}
 & Ours & \textbf{0.9278} &\textbf{0.9318} & \textbf{0.9313} & \textbf{0.9109}& \textbf{0.8914} & \textbf{0.8983} & \textbf{0.8959}& \textbf{0.8900} \\ \cline{2-10}
 \hline
\end{tabular}
\label{tab: tab12}
\end{table*}

\begin{figure*}[!t]
\centering
\subfigure{
\begin{minipage}[b]{0.99\textwidth}
\includegraphics[trim={1.88cm 0 0 0},clip,width=0.24\textwidth]{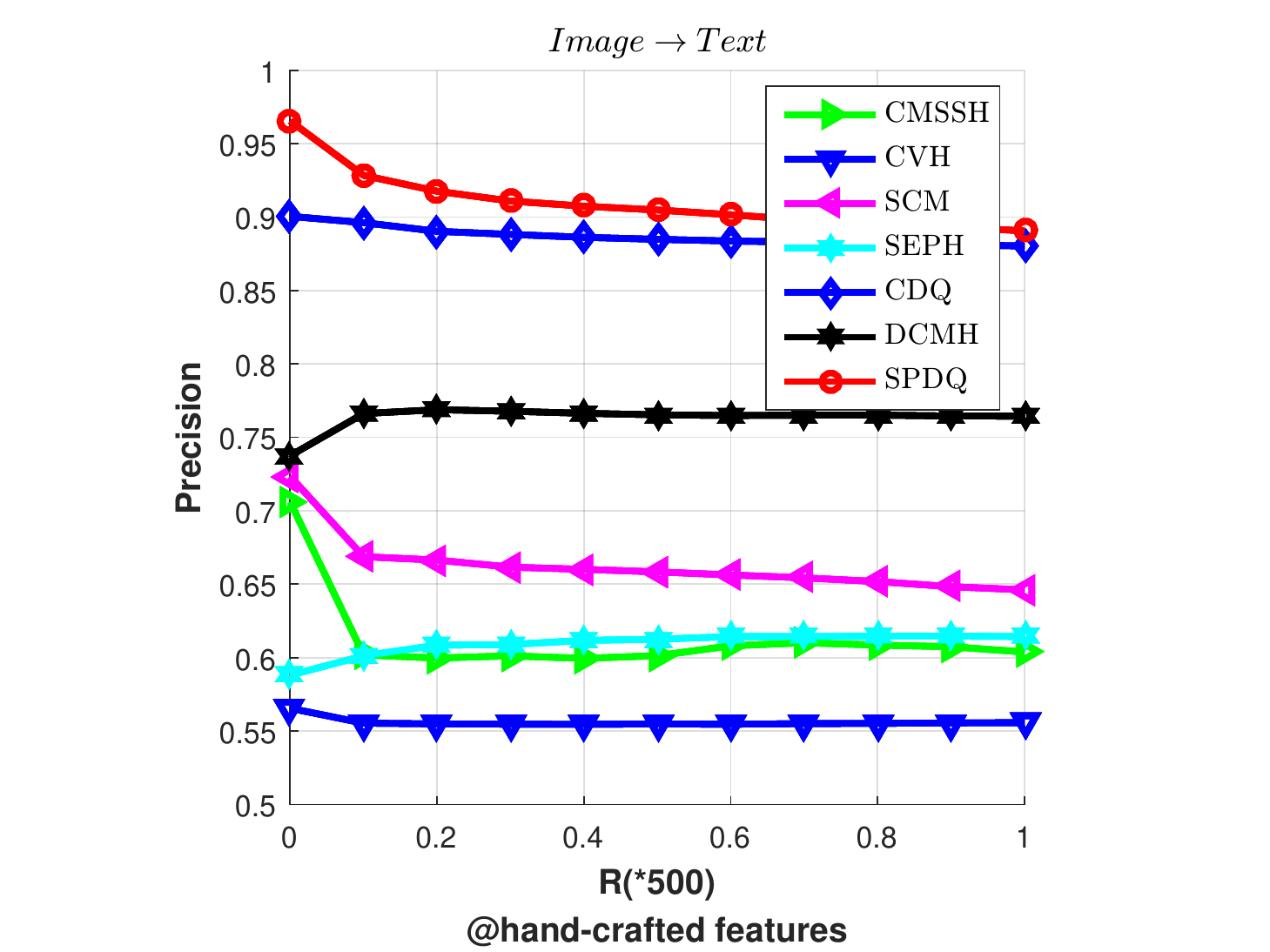}
\includegraphics[trim={1.88cm 0 0 0},clip,width=0.24\textwidth]{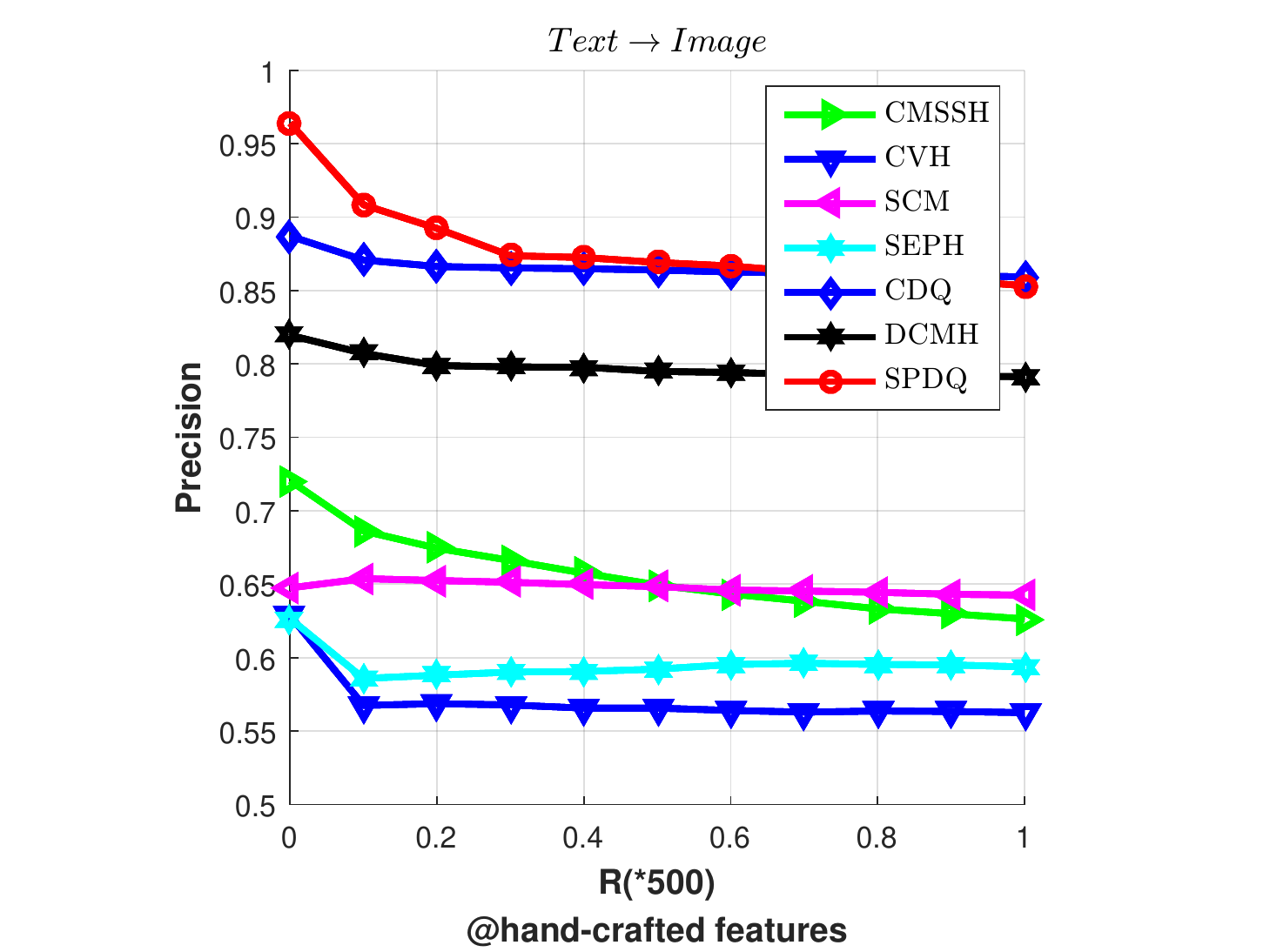}
\includegraphics[trim={1.88cm 0 0 0},clip,width=0.24\textwidth]{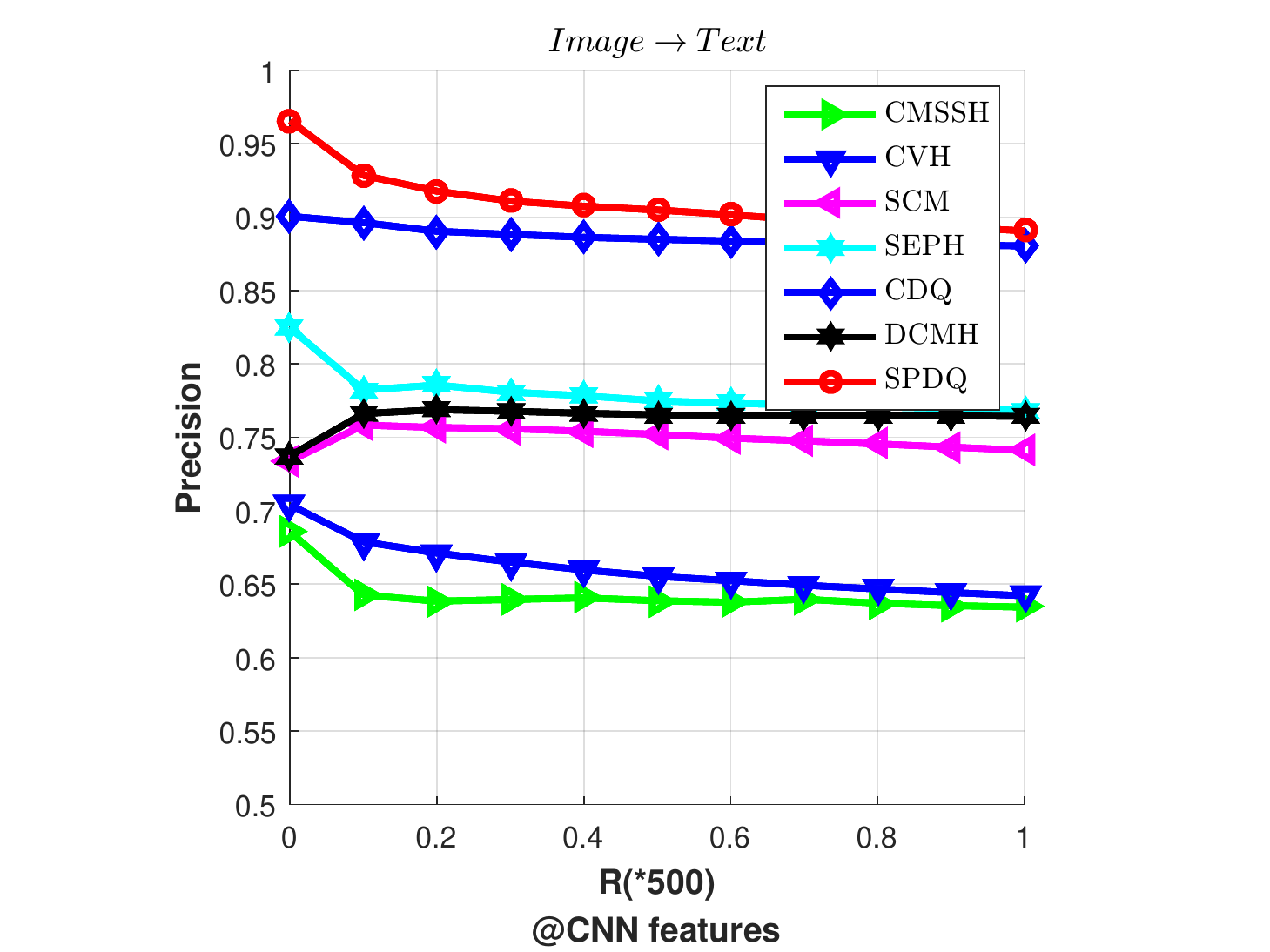}
\includegraphics[trim={1.88cm 0 0 0},clip,width=0.24\textwidth]{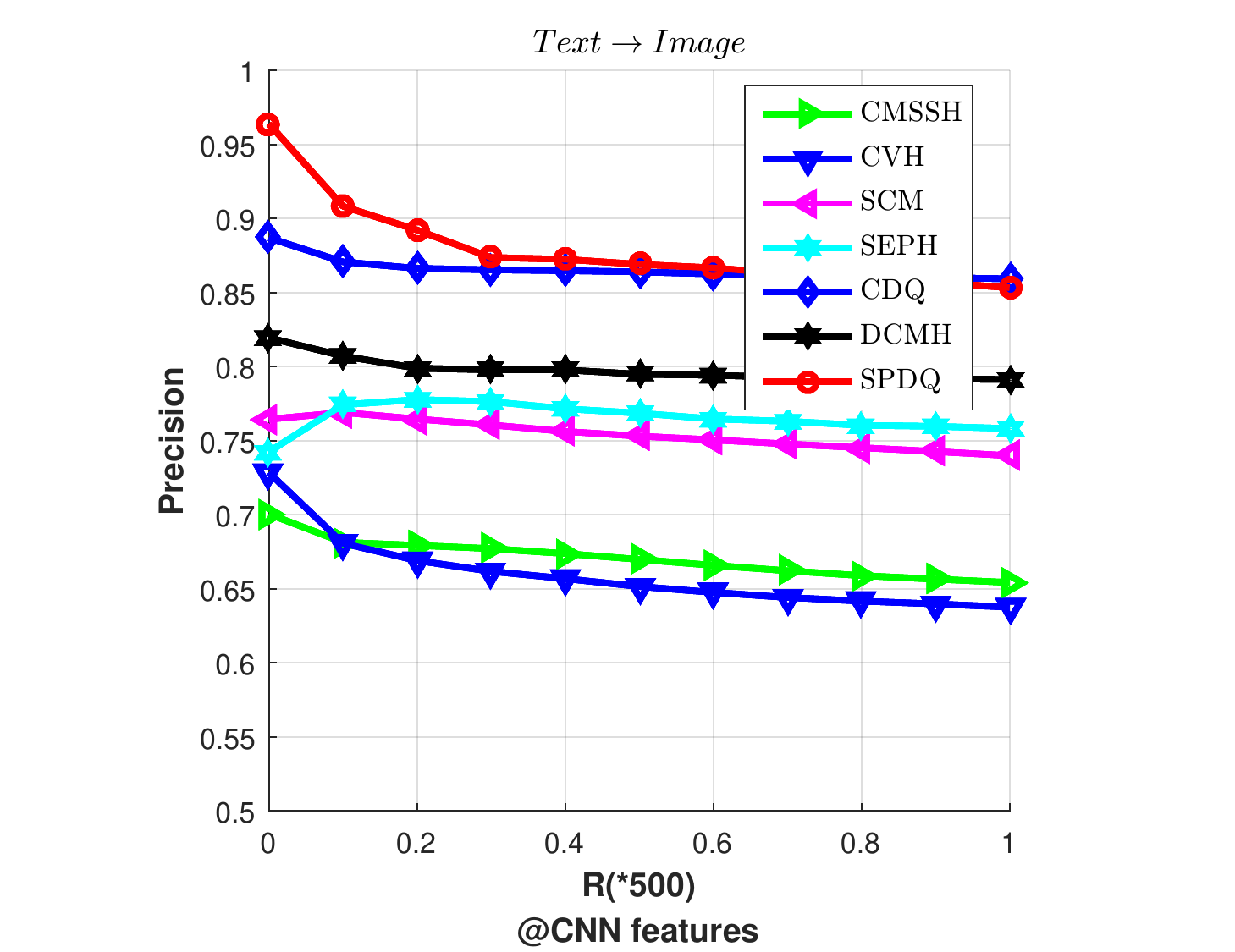}
\end{minipage}
}
%\vspace{-0.3cm}
\caption{Precision values at various numbers of top retrieved data points on FLICKR25K with code length 16.}
\label{fig: top_1}
\end{figure*}
\begin{figure*}[!t]
\centering
\subfigure{
\begin{minipage}[b]{0.99\textwidth}
\includegraphics[trim={1.88cm 0 0 0},clip,width=0.24\textwidth]{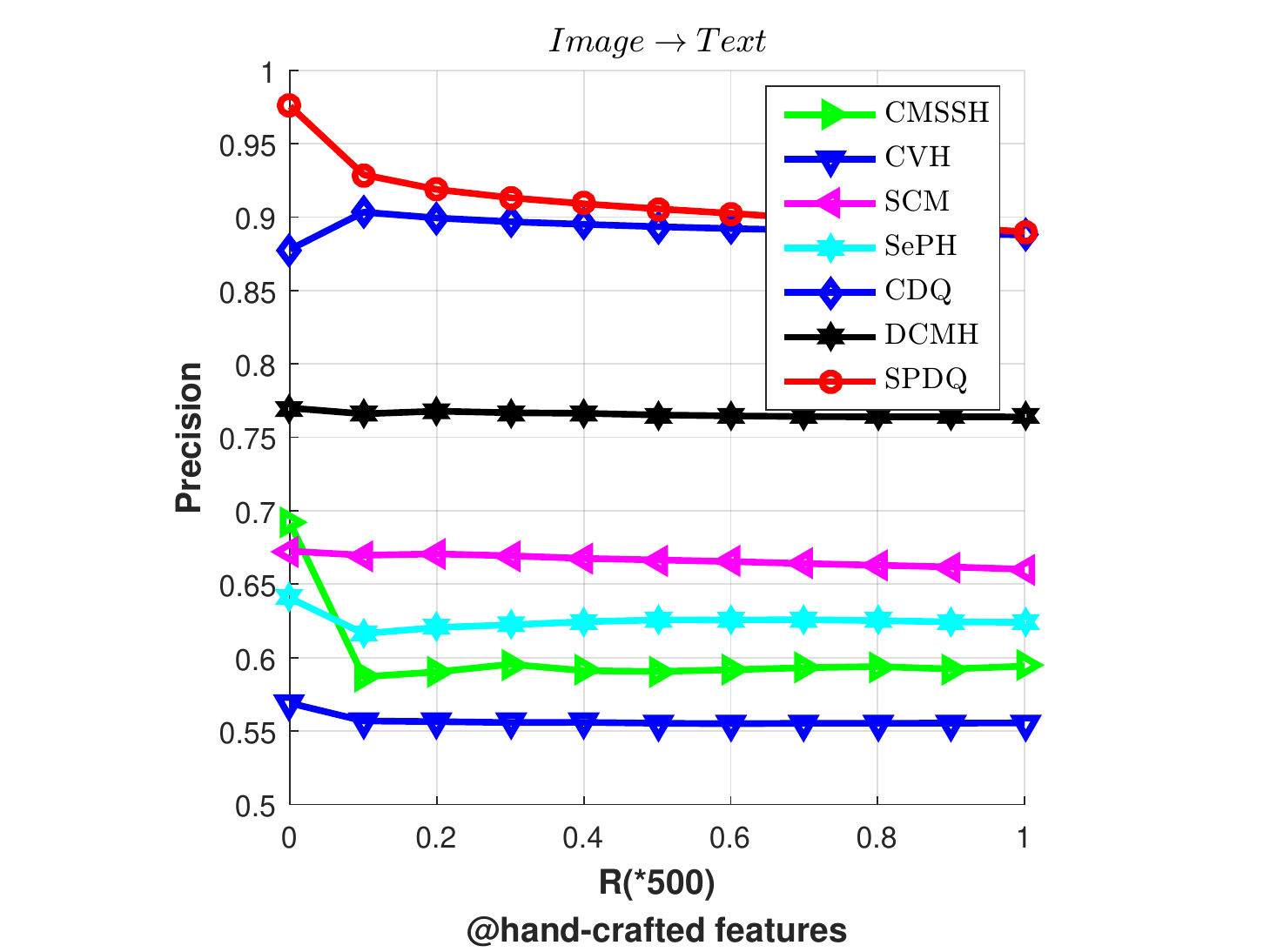}
\includegraphics[trim={1.88cm 0 0 0},clip,width=0.24\textwidth]{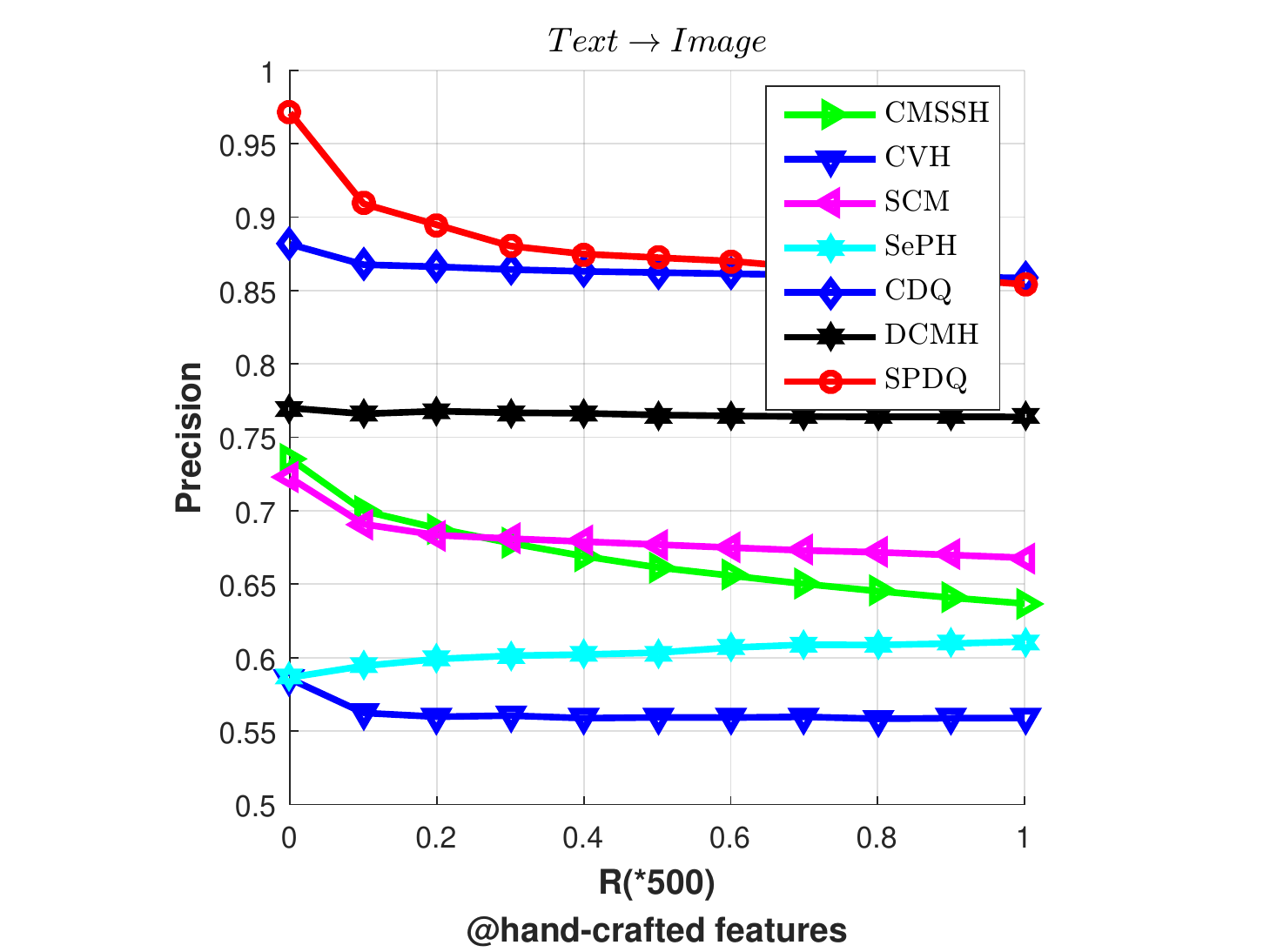}
\includegraphics[trim={1.88cm 0 0 0},clip,width=0.24\textwidth]{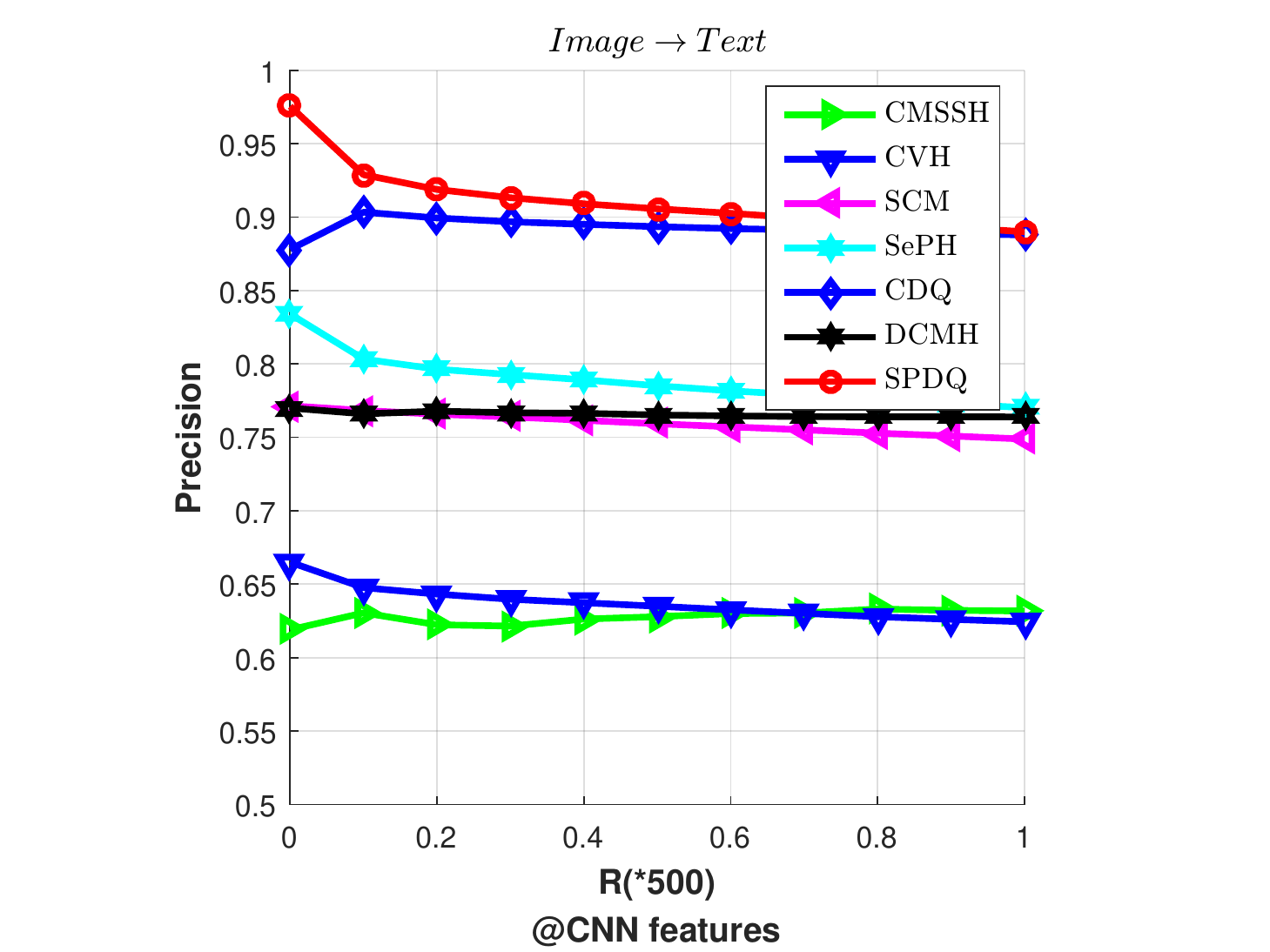}
\includegraphics[trim={1.88cm 0 0 0},clip,width=0.24\textwidth]{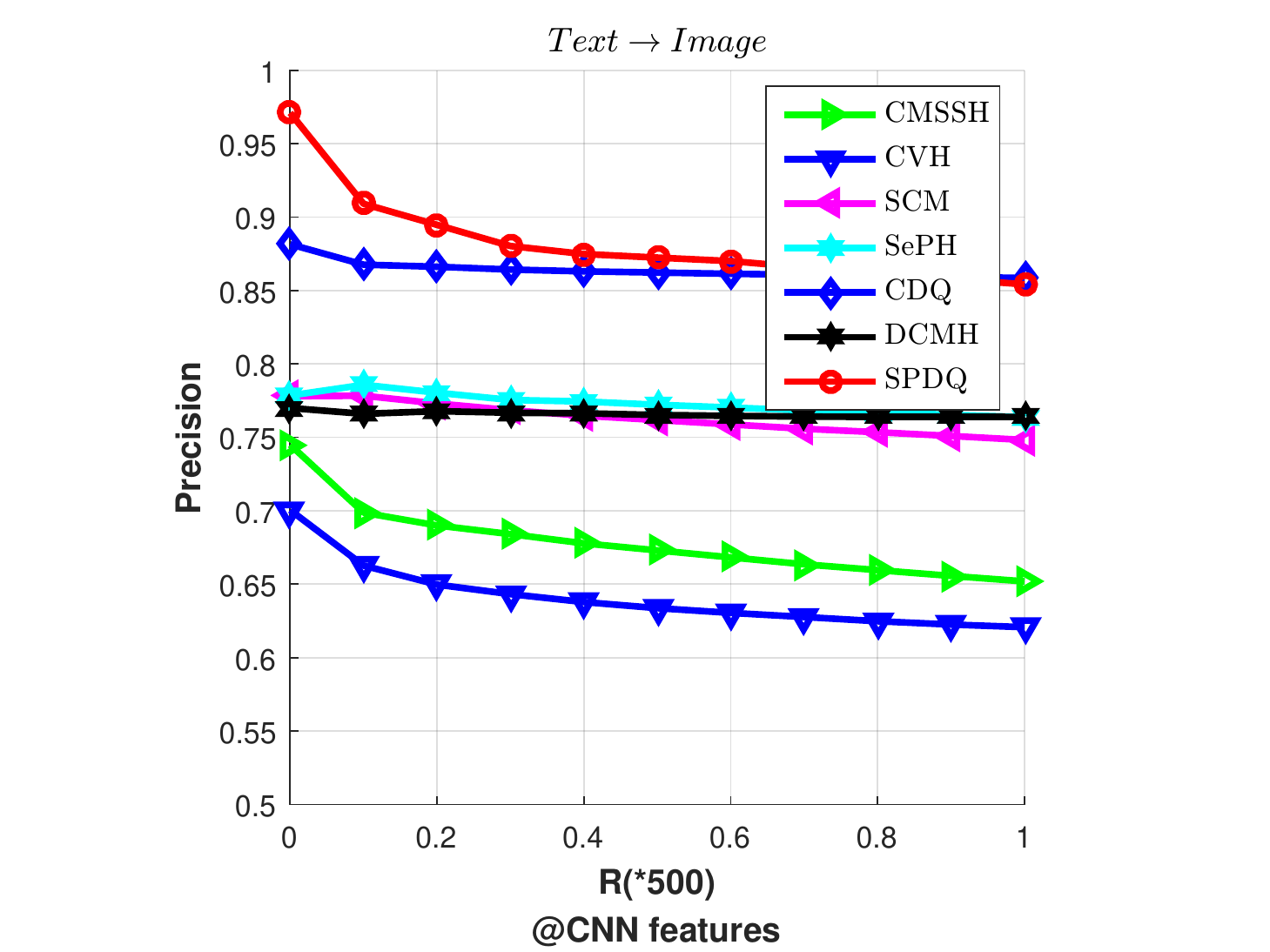}
\end{minipage}
}
%\vspace{-0.3cm}
\caption{Precision values at various numbers of top retrieved data points on FLICKR25K with code length 32.}
\label{fig: top_2}
\end{figure*}
\begin{figure*}[!t]
\centering
\subfigure{
\begin{minipage}[b]{0.99\textwidth}
\includegraphics[trim={1.88cm 0 0 0},clip,width=0.24\textwidth]{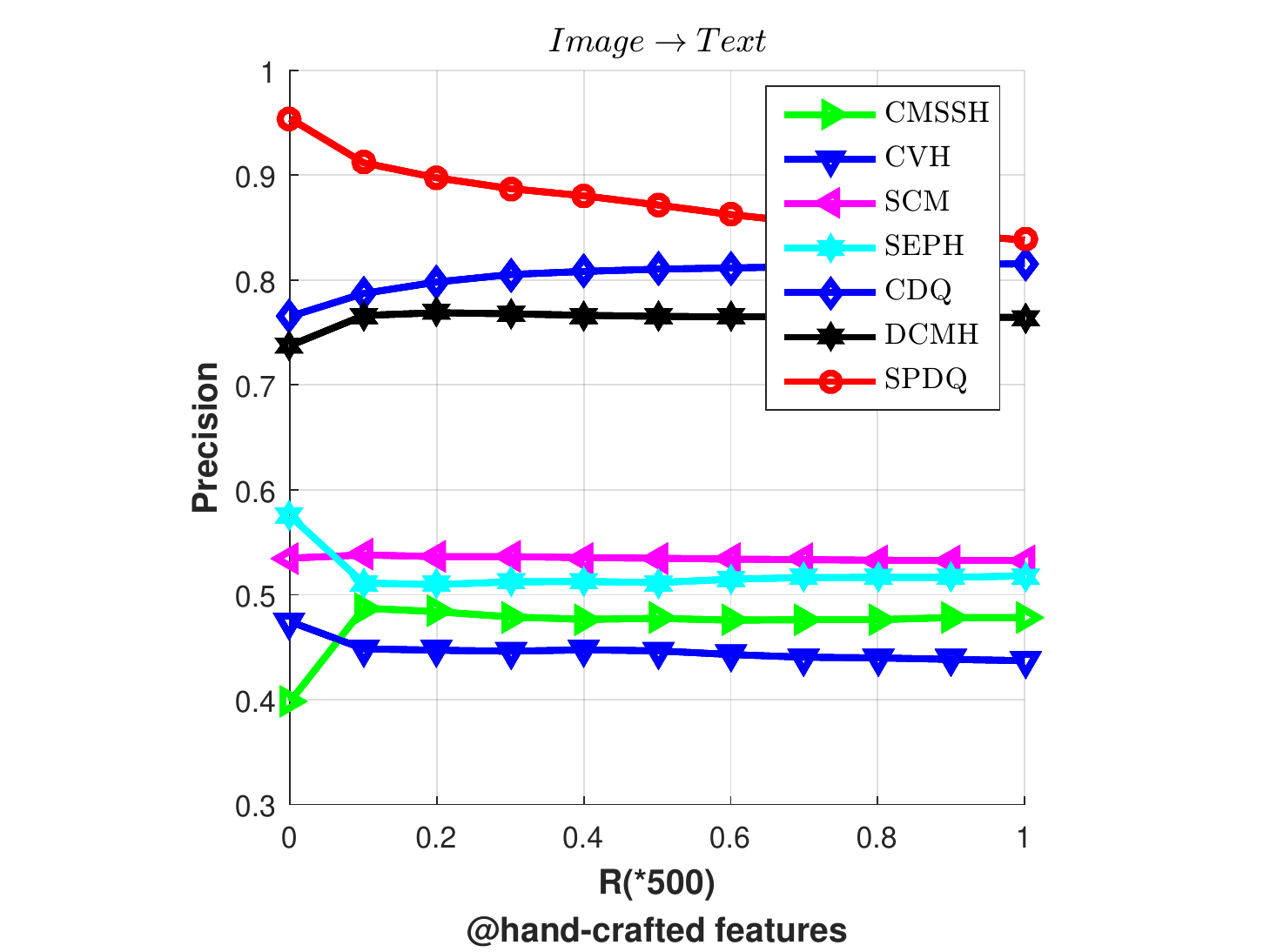}
\includegraphics[trim={1.88cm 0 0 0},clip,width=0.24\textwidth]{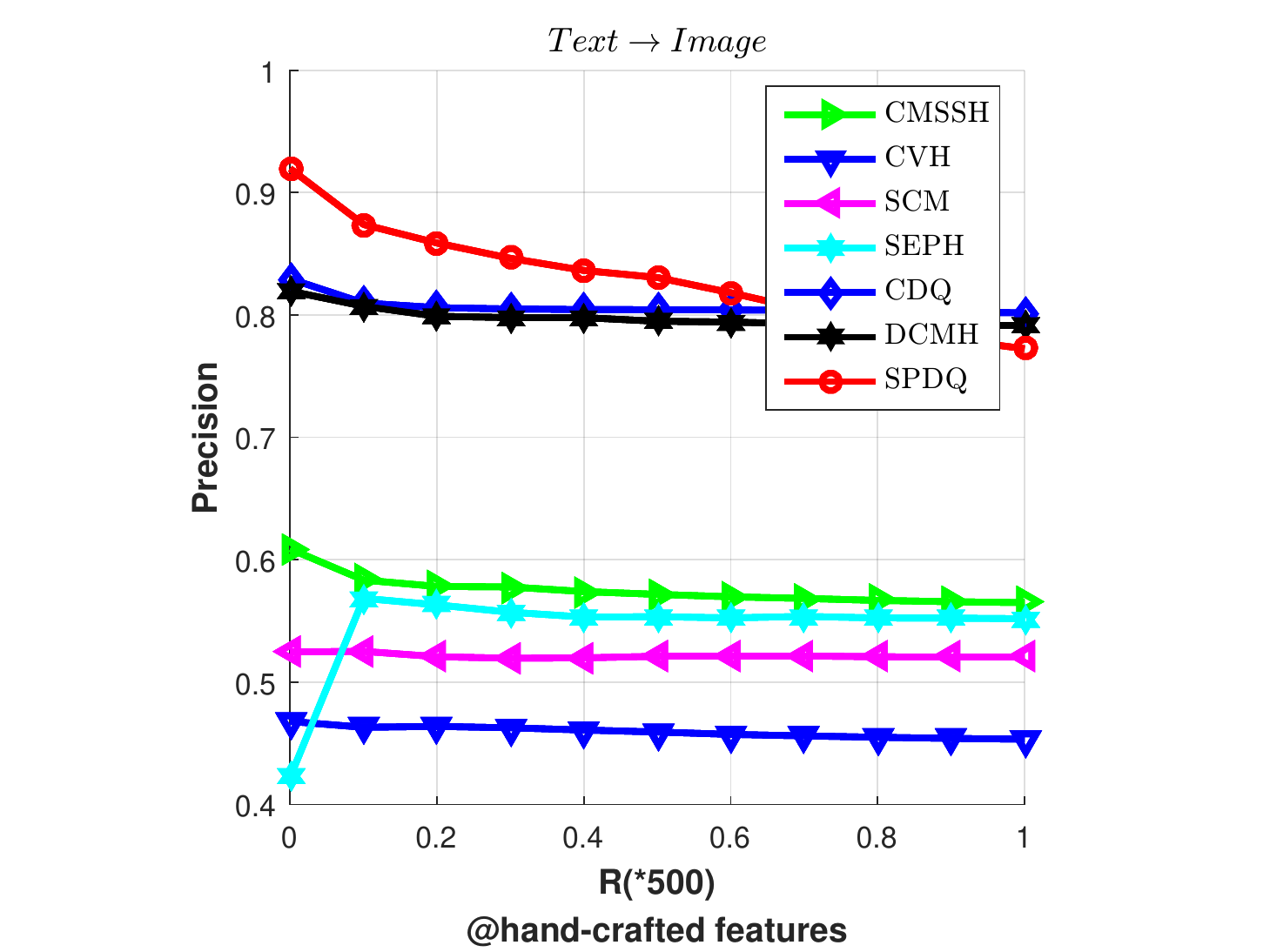}
\includegraphics[trim={1.88cm 0 0 0},clip,width=0.24\textwidth]{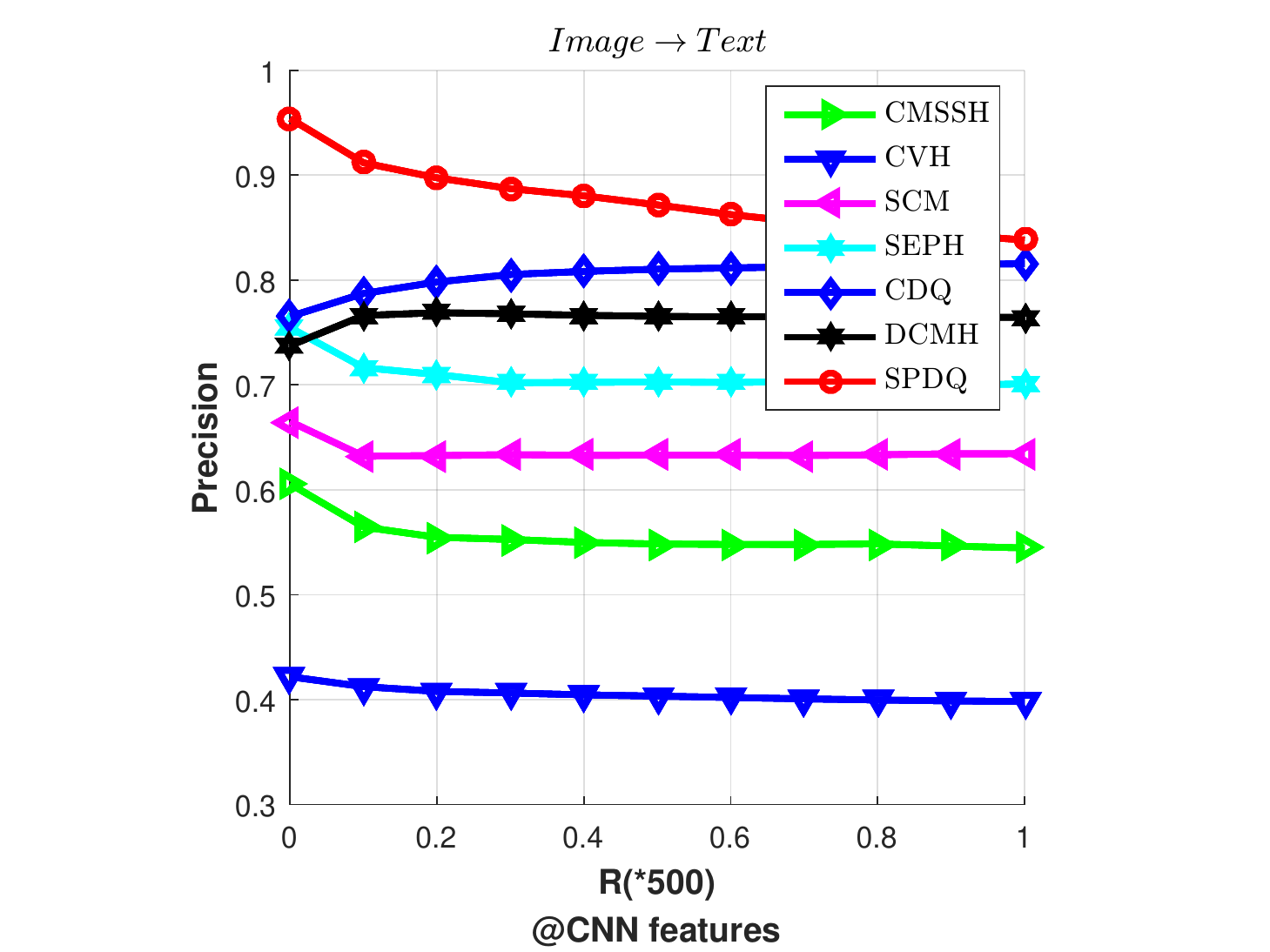}
\includegraphics[trim={1.88cm 0 0 0},clip,width=0.24\textwidth]{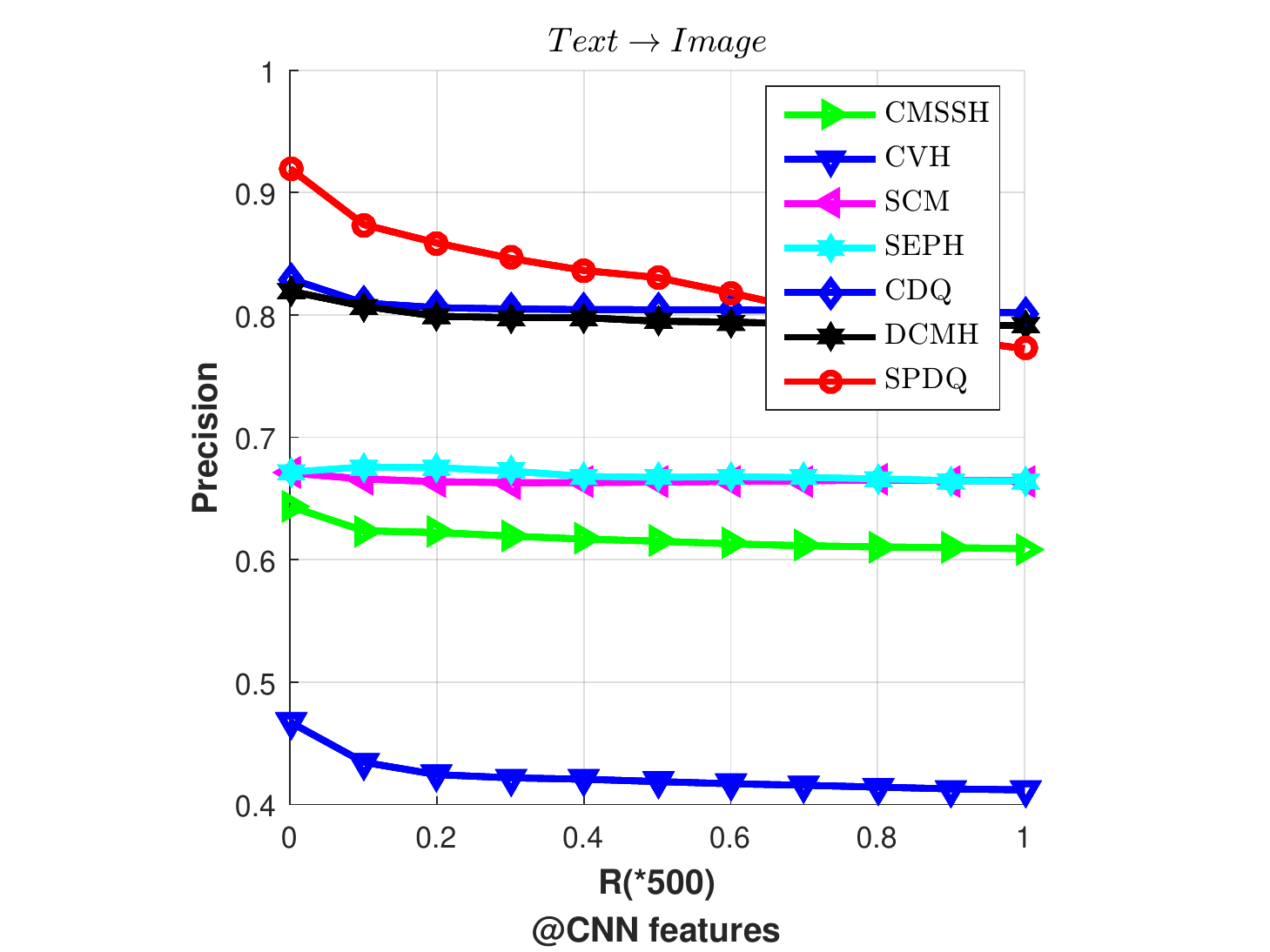}
\end{minipage}
}
%\vspace{-0.3cm}
\caption{Precision values at various numbers of top retrieved data points on NUS-WIDE with code length 16.}
\label{fig: top_3}
\end{figure*}
\begin{figure*}[!t]
\centering
\subfigure{
\begin{minipage}[b]{0.99\textwidth}
\includegraphics[trim={1.88cm 0 0 0},clip,width=0.24\textwidth]{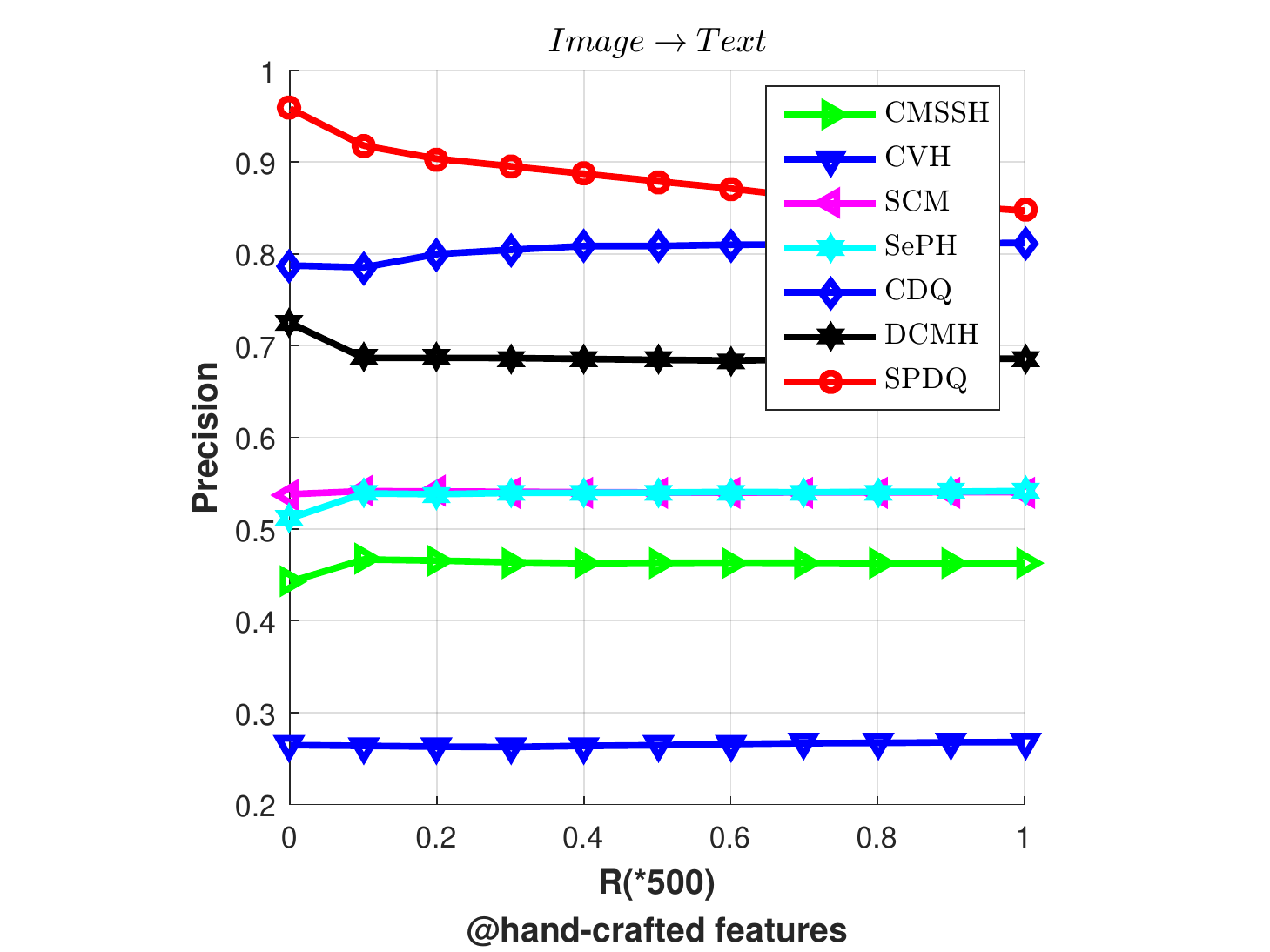}
\includegraphics[trim={1.88cm 0 0 0},clip,width=0.24\textwidth]{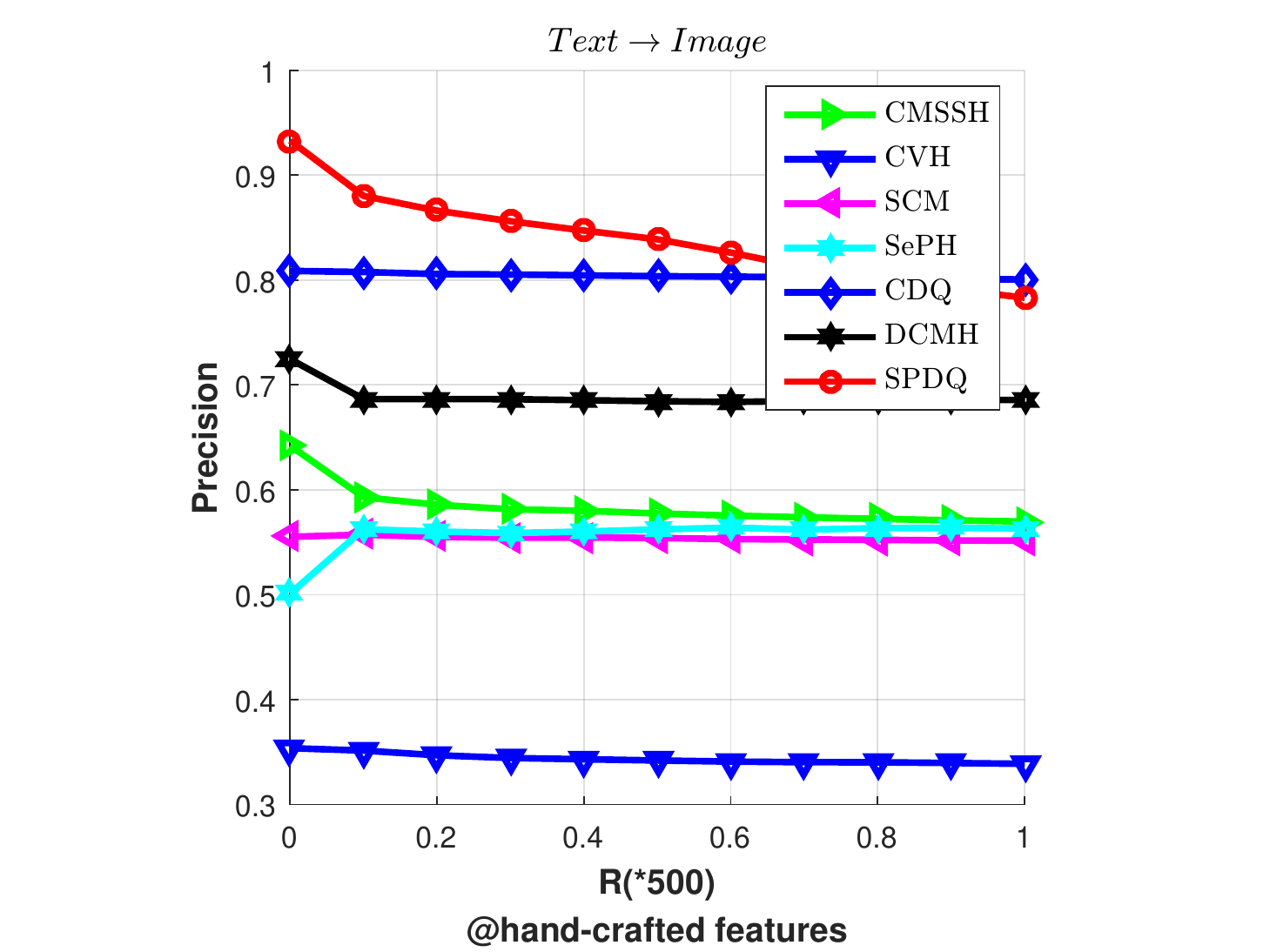}
\includegraphics[trim={1.88cm 0 0 0},clip,width=0.24\textwidth]{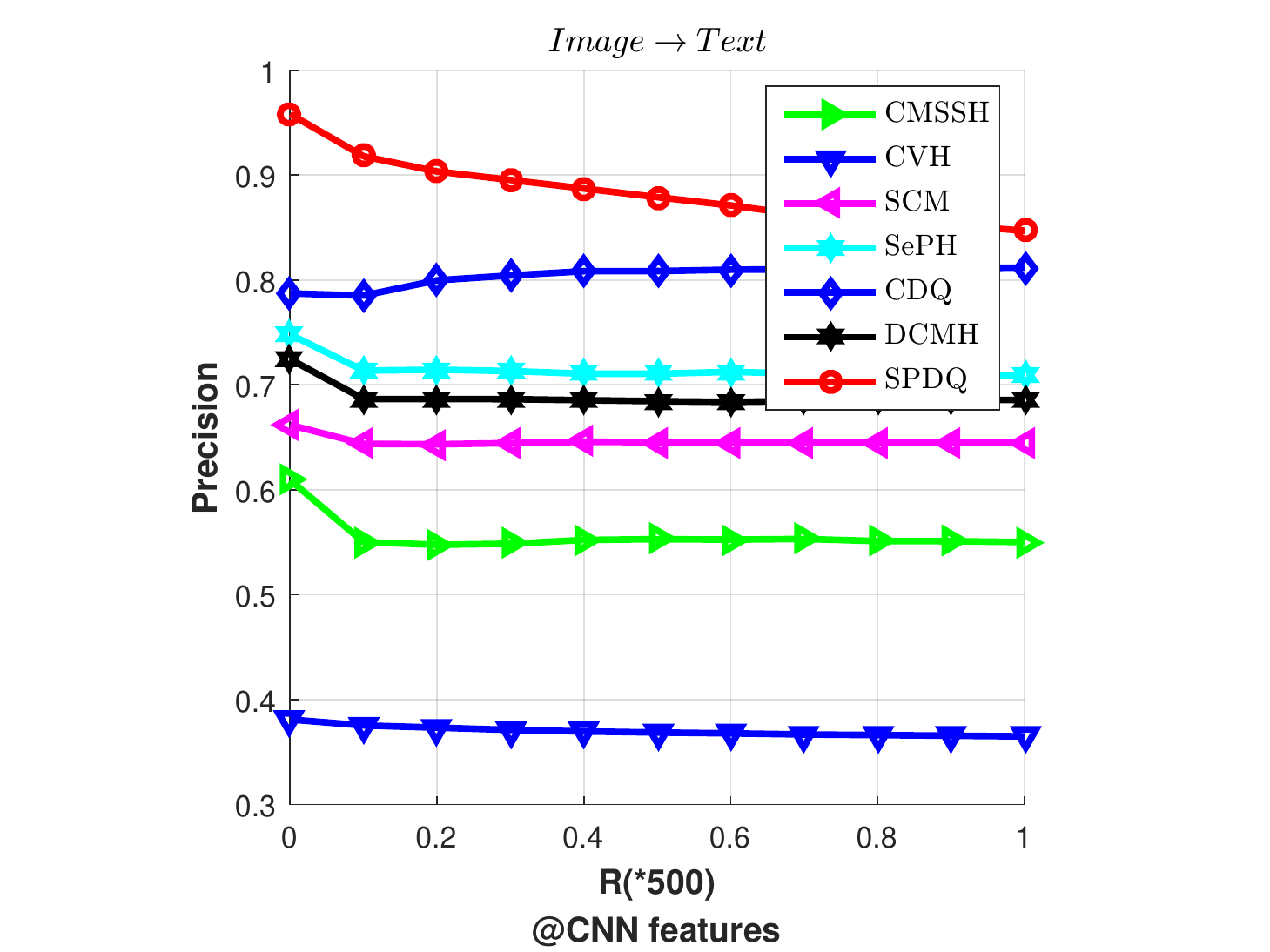}
\includegraphics[trim={1.88cm 0 0 0},clip,width=0.24\textwidth]{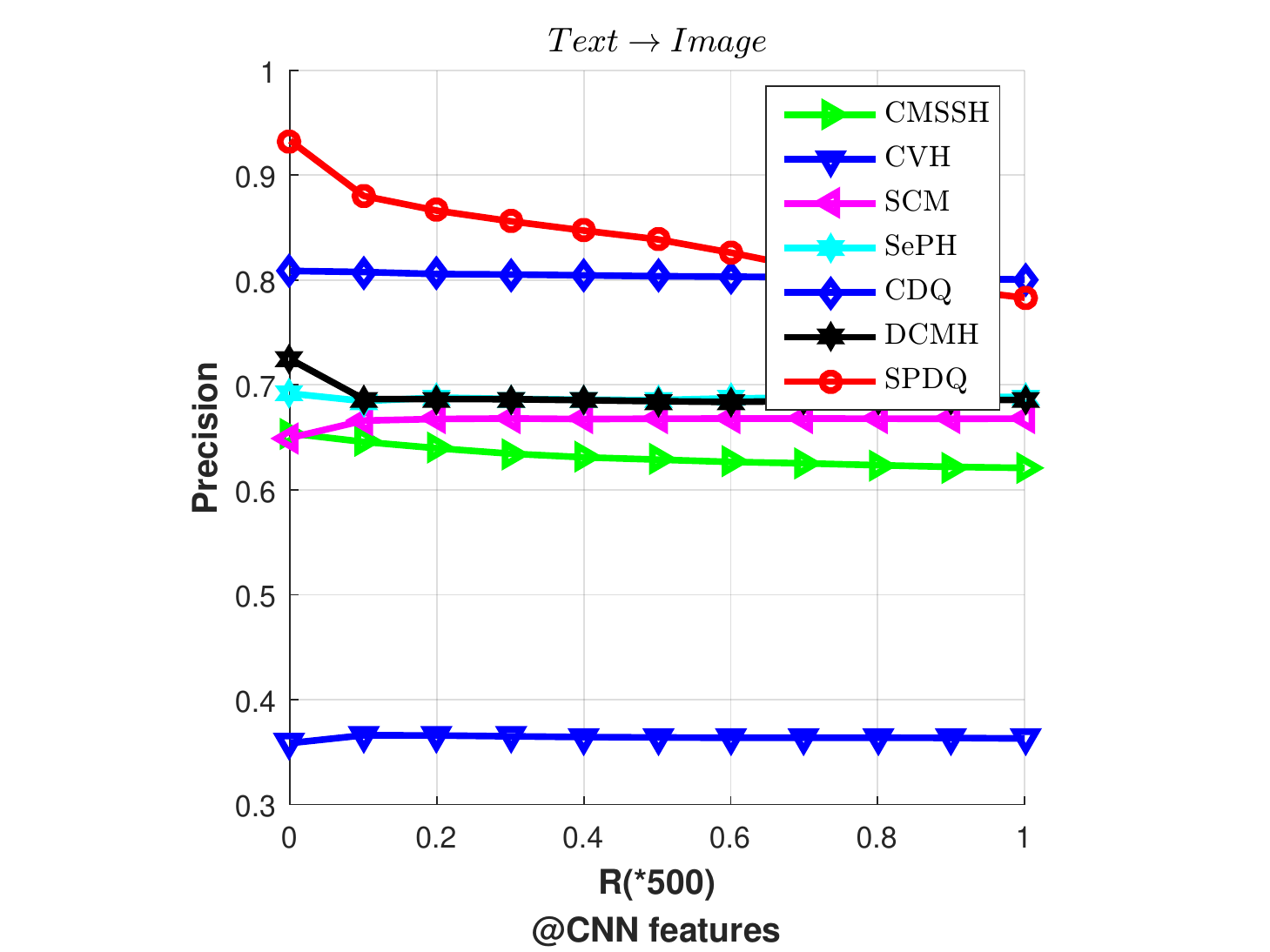}
\end{minipage}
}
%\vspace{-0.3cm}
\caption{Precision values at various numbers of top retrieved data points on NUS-WIDE with code length 32.}
\label{fig: top_4}
\end{figure*}

\subsection{Setup}
\textbf{Datasets.} We evaluate our method on two popular benchmark datasets, \textbf{FLICKR25K} and \textbf{NUS-WIDE}. All the data sets have two modalities, i.e. image
and text. Some statistics of them are introduced below.

\textbf{FLICKR25K} originally contains 25,000 images collected from the Flickr website. Each image associates with several textual tags and is manually annotated with at least one of 24 provided unique labels. In our experiment, we only keep those textual tags appearing at least 20 times and remove image-tag pairs without manually annotated labels. Then we get 20,015 image-tag pairs for experiment. The text for each point is represented as a 1,386-dimensional bag-of-words vector. The hand-crafted feature for each image is represented by a 512-dimensional GIST feature
vector.

\textbf{NUS-WIDE} contains 269,648 images with associated tags, where each pair is annotated with multiple labels among 81 concepts. Following prior works~\cite{ding2014collective,zhen2012co}, we use the subset of 186,577 image-text pairs that belong to 10 most popular concepts.  The text for each point is represented as a 1,000-dimensional bag-of-words vector. The hand-crafted feature for each image is a 500-dimensional bag-of-visual
words (BOVW) vector.

For our proposed approach and deep learning based compared methods, we directly use raw pixels as the image modality inputs. For traditional shallow methods, besides the shallow features introduced above, we extract 4096-dimensional feature vectors from the last fully connected layer by using AlexNet architecture~\cite{krizhevsky2012imagenet} pre-trained on ImageNet dataset. So, we compared our method with traditional shallow methods with hand-crafted features as well as deep features.  For FLICKR25K, we randomly select 2,000 instances as a test set, the rest are used as a validation set, from which we randomly select 10,000 instances as a training set.  For NUS-WIDE, following~\cite{wang2016multimodal}, we randomly select 1866 of the dataset as a test set, the rest are used as a validation set, from which we randomly select 10,000 instances as a training set.

\textbf{Evaluation.}
Two evaluation criteria are adopted to evaluate the performance of the proposed method, namely mean of average precision (MAP) and topN-precision. These two criteria are based on Hamming ranking which ranks all the data points based on the Hamming distances to the query.

MAP is one of the most widely-used criteria to evaluate retrieval accuracy. Given a query and a list of $R$ ranked retrieval results, the average precision (AP) for this query is defined as
\begin{equation}
\begin{aligned}
AP = \frac{1}{N}\mathop{\sum}_{r=1}^{R}P(r)\delta{(r)},
\end{aligned}
\label{eqn: eq12}
\end{equation}
where $N$ in the number of ground-truth relevant instances in the database for the query, and $P(r)$ presents the precision for the top $r$ retrieved instances. $\delta{(r)} = 1$ when the $r$-th retrieval instance is relevant to the query, otherwise $\delta{(r)} = 0$. MAP is defined as the average of APs for all the queries. $R$ is set to 50 in our experiments. The ground-truth relevant instances for a query are defined as those sharing at least one label with it. TopN-precision shows the precision at different numbers of retrieved instances.

\textbf{Compared methods.}
We compare the proposed approach with various state-of-the-art cross-modal similarity search methods. Specifically, we take six supervised methods: SCM\cite{zhang2014large}, SePH\cite{lin2015semantics}, CMSSH~\cite{bronstein2010data}, CVH~\cite{kumar2011learning}, DCMH~\cite{jiang2016deep}, CDQ~\cite{cao2017collective}, as baselines, where DCMH is deep hashing cross-modal method, and CDQ is deep quantization cross-modal method. Source codes of all baseline methods are kindly provided by the authors. Since, SePH is a kernel-based method, we use RBF kernel and take 500 randomly selected points as kernel bases by following its authors' suggestion. The parameters for all the above methods are set according to the original papers.

\textbf{Implementation details.}
As shown in Fig.~\ref{fig: model}, the hybrid deep architecture constitutes an image network and a text network. For the image network, we adopt the AlexNet architecture~\cite{krizhevsky2012imagenet}, and replace the last fully connected layer by a shared layer and a private layer. We set the shared layer with 256 units and the private layer with 48 units and use the hyperbolic tangent (tanh)
function as the activation functions. Parameters for the shared layers and the private layers are learned from scratch, and all parameters for the preceding layers are fine-tuned from the model pre-trained on ImageNet dataset. For text modality, we use a three-layer MLP, where the first and the second fully connected layers both have 4,096 units, and the last fully connected layer is replaced by a shared layer and a private layer, which have the same number units as in the image network. All of the parameters for the text network are learned from scratch. We employ the standard stochastic gradient descent algorithm for optimization with 0.9 momentum, fix min-batch size to 128, and the learning rate is chosen from $10^{-6}$ to $10^{-1}$ with a validation set. All experiments are run for five times, and the average performance is reported. Similarly to~\cite{long2016composite}, we set the number of elements in each dictionary as $K=256$, so for each data point, the binary code length $M = C{\log}_{2}D = 8C$ bits. We can set $C = {M}\verb|/|{8}$ when $M$ is known.

\subsection{Results and Discussions}
For all the datasets, we first present MAP values of all the methods for various code lengths to providing a global evaluation. Then we report the topN-precision curves with code length 16 and 32 to make a comprehensive contrastive study.

The MAP results for SPDQ, DCMH, CDQ and other baselines with hand-crafted features on FLICKR25K and NUS-WIDE datasets are reported in Table~\ref{tab: tab11}. Here, ``$I\to T$''
represents the case where the query is image and the database is text, and ``$T\to I$'' represents the case where the queries are  texts and the database are images. We can find that DCMH, CDQ and SPDQ which are both deep learning based methods outperform all the other shallow baselines with hand-crafted features by a large margin. There may be two possible reasons. The first is that since all these deep learning based methods are build upon some existing deep architectures, and the features extracted from these deep architectures have shown to achieve far superior performance than traditional hand-crafted features in various tasks, so the better performance obtained by these deep learning based methods may be from superior features. The second reason may be that by designing more suitable loss functions or architectures and jointly optimizing the feature extraction and hash code learning, these methods can better capture the semantic information than other methods.

To further verify the effectiveness of our SPDQ, for image modality, we extract CNN features from AlexNet pre-trained on ImageNet dataset, which is the same as the initial CNN for image modality in SPDQ. Then, all the shallow
baselines are trained based on these CNN features. Table~\ref{tab: tab12} shows the MAP values for all the methods with code length varying from 16-bit to 128-bit. Compared Table~\ref{tab: tab11} and Table~\ref{tab: tab12}, we can see that most of the shallow methods obtain an increase in MAP values in all code lengths, which demonstrates that features extracted from existing deep architectures can usually preserve more semantic information and achieve better performance than traditional hand-crafted features. Despite that, SPDQ and CDQ usually outperform other baseline methods, and DCMH achieves comparable results in many cases. One of the reasons may be that by incorporating feature learning and hash code learning in an end-to-end architecture, these methods can jointly optimize these two parts thus causing better results. Besides, as shown in Table~\ref{tab: tab12}, the proposed SPDQ outperforms all  shallow baselines using CNN features with different code lengths on all retrieval tasks, and also outperforms DCMH, the state-of-the-art deep cross-modal hashing method, as well as CDQ the state-of-the-art deep cross-modal quantization method, which  well demonstrates the superiority of the proposed method.

TopN-precision curves on FLICKR25K and NUS-WIDE are showed in Fig.~\ref{fig: top_1}, Fig.~\ref{fig: top_2}, Fig.~\ref{fig: top_3}, and Fig.~\ref{fig: top_4}, where Fig.~\ref{fig: top_1} and Fig.~\ref{fig: top_2} show the result with code length 16, and Fig.~\ref{fig: top_3} and Fig.~\ref{fig: top_4} show the results with code length 32. For each figure, the first two sub-figures are based on hand-crafted features and the last two sub-figures are based on CNN features for shallow baseline methods. It can be seen that the proposed approach outperforms all the baseline methods for both hand-crafted features and CNN features. Our SPDQ can also achieve the best performance on other cases with different code lengths, such as 64 and 128. Those results are omitted due to space limitation. The result of TopN-precision curves are consistent with the mAP evaluation. In retrieval system, since the users always focus on the front returned results, the relevance of the top returned instances with the query is usually more important. From the topN-precision curves, we can find that the proposed approach usually outperforms other methods by a large margin when the number of the returned instances are relatively small.

\subsection{Effect of Training Size}
\begin{figure}
	\centering
	\subfigure{
		\begin{minipage}[b]{0.98\textwidth}
			\includegraphics[width=0.24\textwidth]{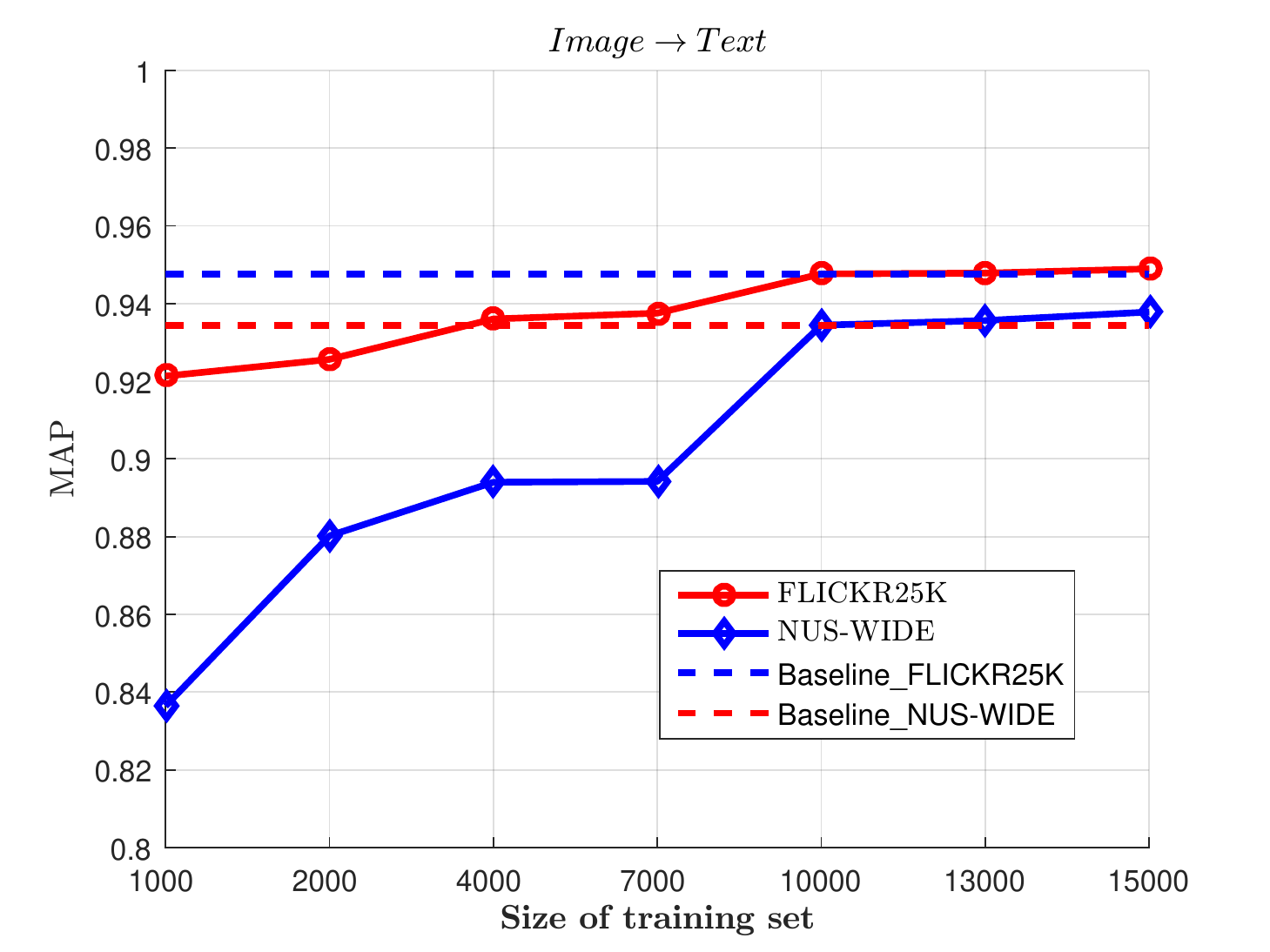}
			\includegraphics[width=0.24\textwidth]{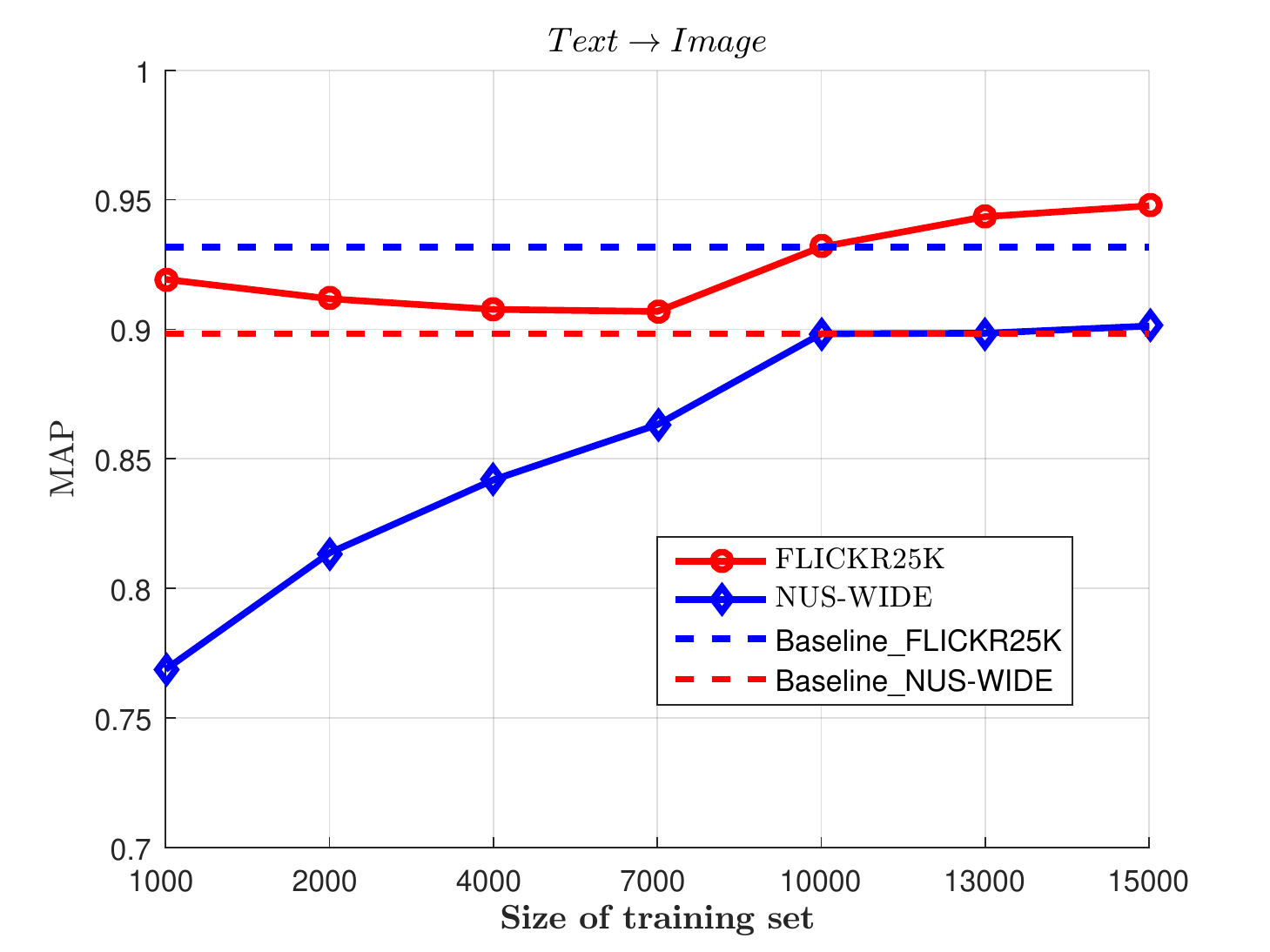}
		\end{minipage}
	}
	%\vspace{-0.3cm}
	\caption{MAP versus the size of training set.}
	\label{fig: train_size}
\end{figure}
In this subsection, we analyze the effects of training set size on the performance of SPDQ on FLICKR25K and NUS-WIDE dataset. Specifically, we fix the hash code length at 32 bits, and vary the training set size from 1,000 to 15,000. Then we measure MAP values of SPDQ and show the results in Fig.~\ref{fig: train_size}. MAP values with training set size 10,000 are chosen as the baseline values. It is easy to see that SPDQ usually obtain better retrieval performance with larger training set size. The increasing trend is particularly apparent for NUS-WIDE dataset, which may be because that NUS-WIDE dataset contains far more data points than FLICKR25K and have more diversities, thus needs more training data to capture the relationships of the whole dataset.

\subsection{Parameter Sensitivity Analysis}
 \begin{figure}
	\centering
	\subfigure{
		\begin{minipage}[b]{0.98\textwidth}
			\includegraphics[width=0.24\textwidth]{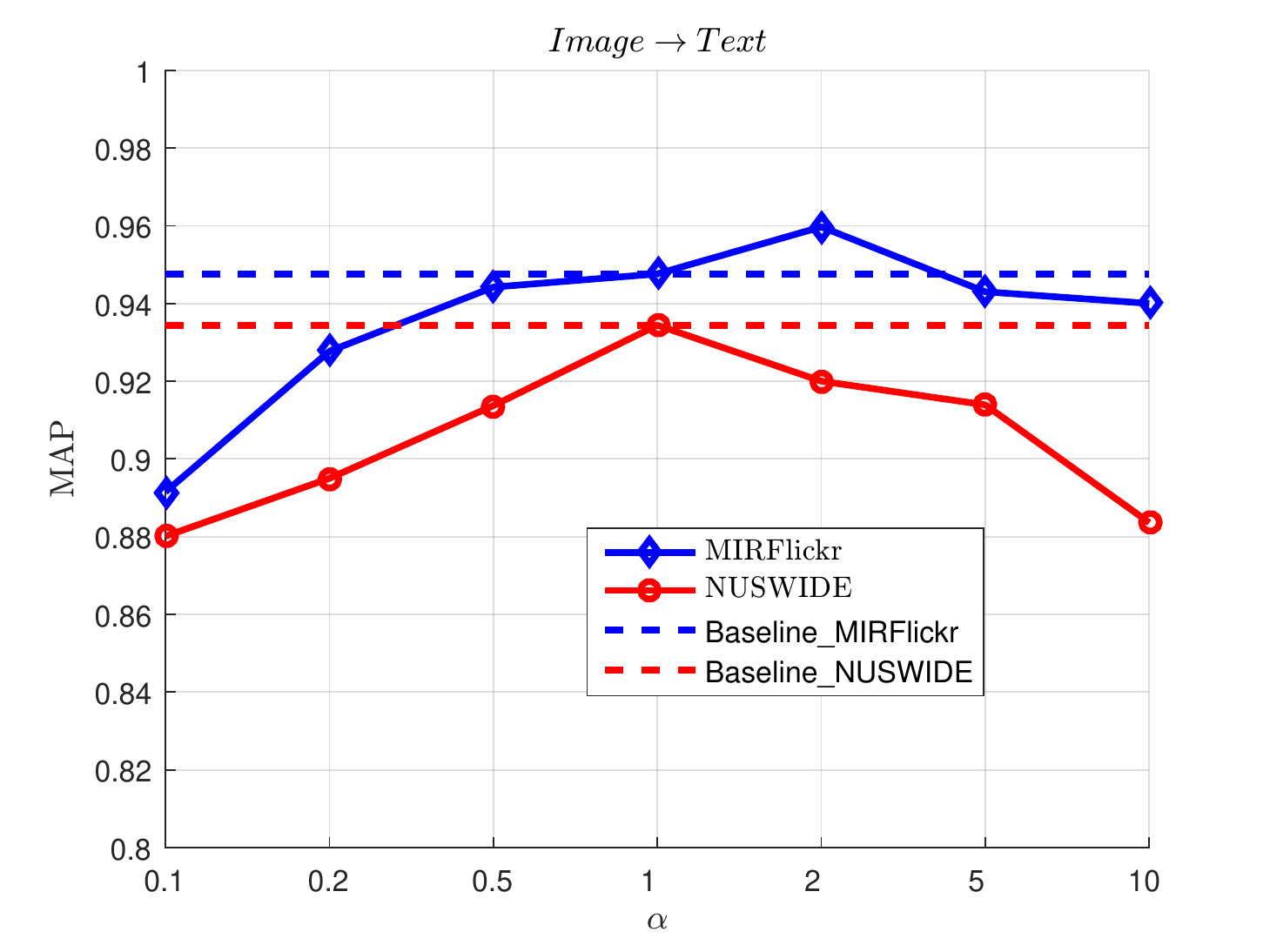}
			\includegraphics[width=0.24\textwidth]{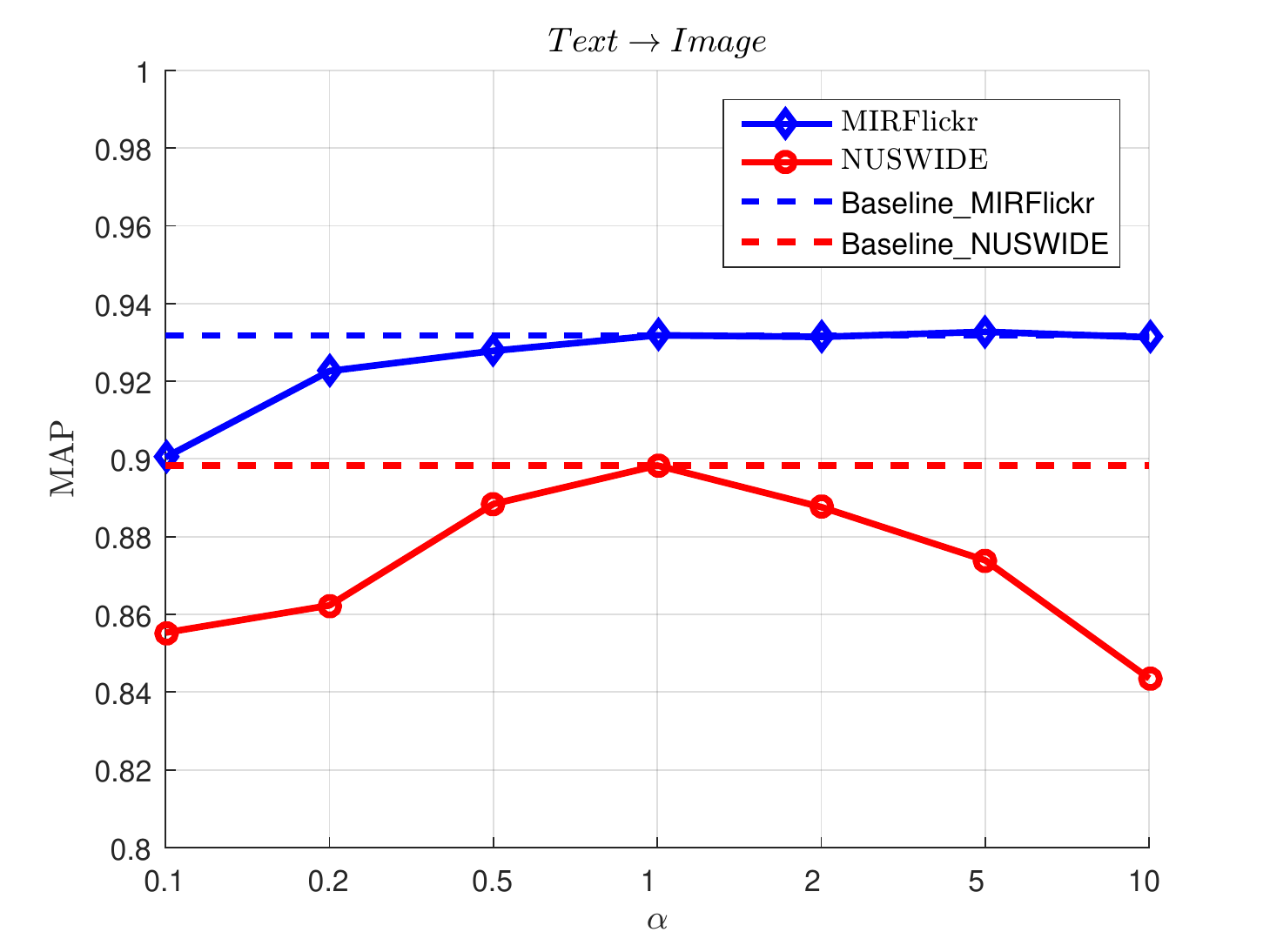}
		\end{minipage}
	}
	%\vspace{-0.3cm}
	\caption{MAP versus parameter $\alpha$.}
	\label{fig: alpha}
\end{figure}
In this part, we evaluate the effects of the parameters of SPDQ. MAP values on the query dataset are reported to study the performance variation with respect to different parameter values. In this experiments, hash code length is fixed as 32 bits, and we conduct the analysis for one parameter by varying its value while fixing the other parameters. The MAP values of SPDQ with $\alpha = 1$, $\beta = 1$, $\lambda = 0.01$ are selected as the baseline values.

The parameter $\alpha$ controls the importance of the classification loss. We vary its value from $0.1$ to $10$. As shown in Fig.~\ref{fig: alpha}, it can be found that the MAP values on both tasks increase with $\alpha$ increasing from $0.1$ to $1$, which may prove that adding the classification loss can enhance the discriminative ability of the learned features. But when $\alpha$ is too large, the classification loss will dominate the training process and effect the quantization learning. In our experiments, we choose $\alpha$ as $1$.

\begin{figure}
	\centering
	\subfigure{
		\begin{minipage}[b]{0.98\textwidth}
			\includegraphics[width=0.24\textwidth]{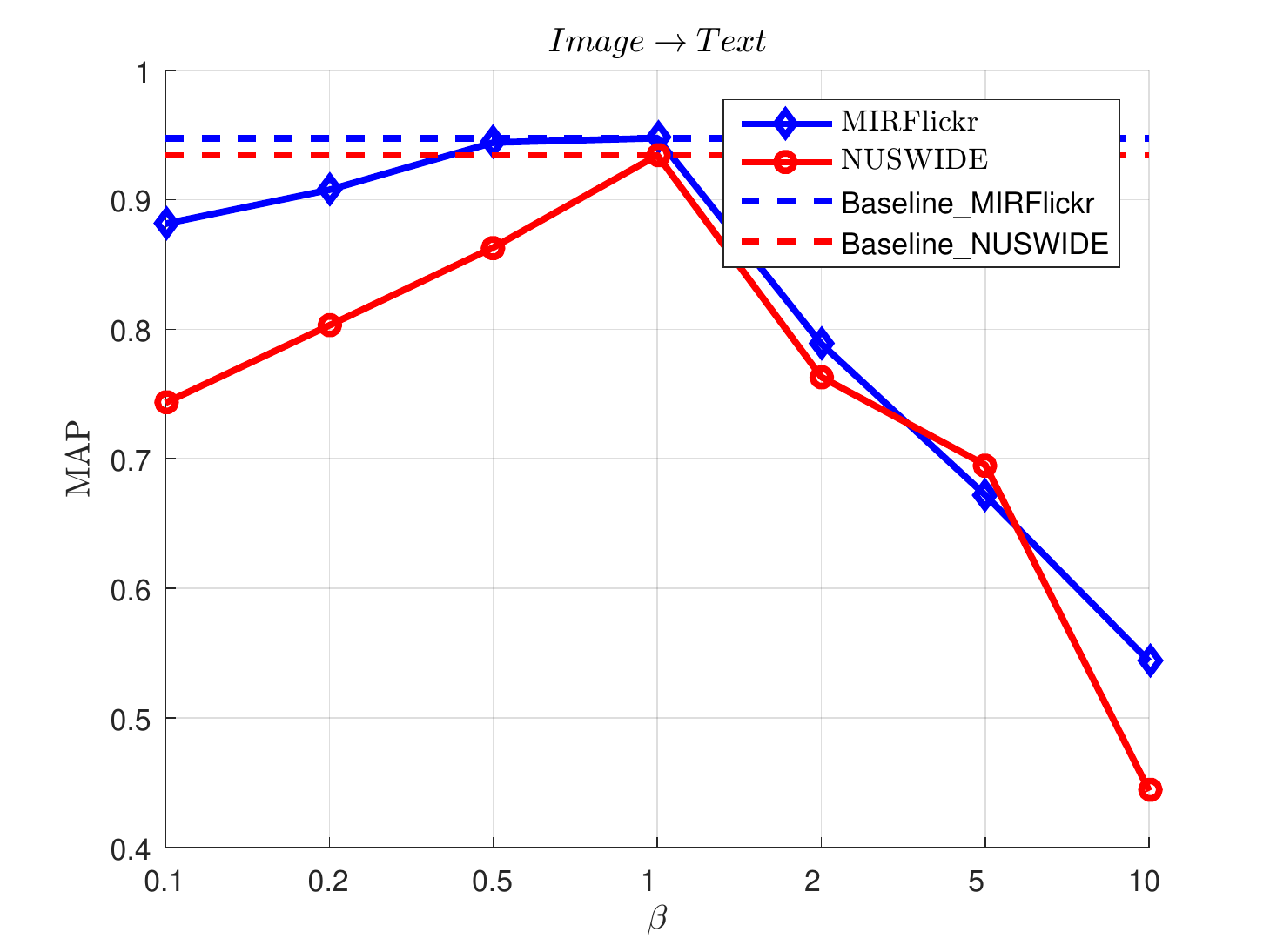}
			\includegraphics[width=0.24\textwidth]{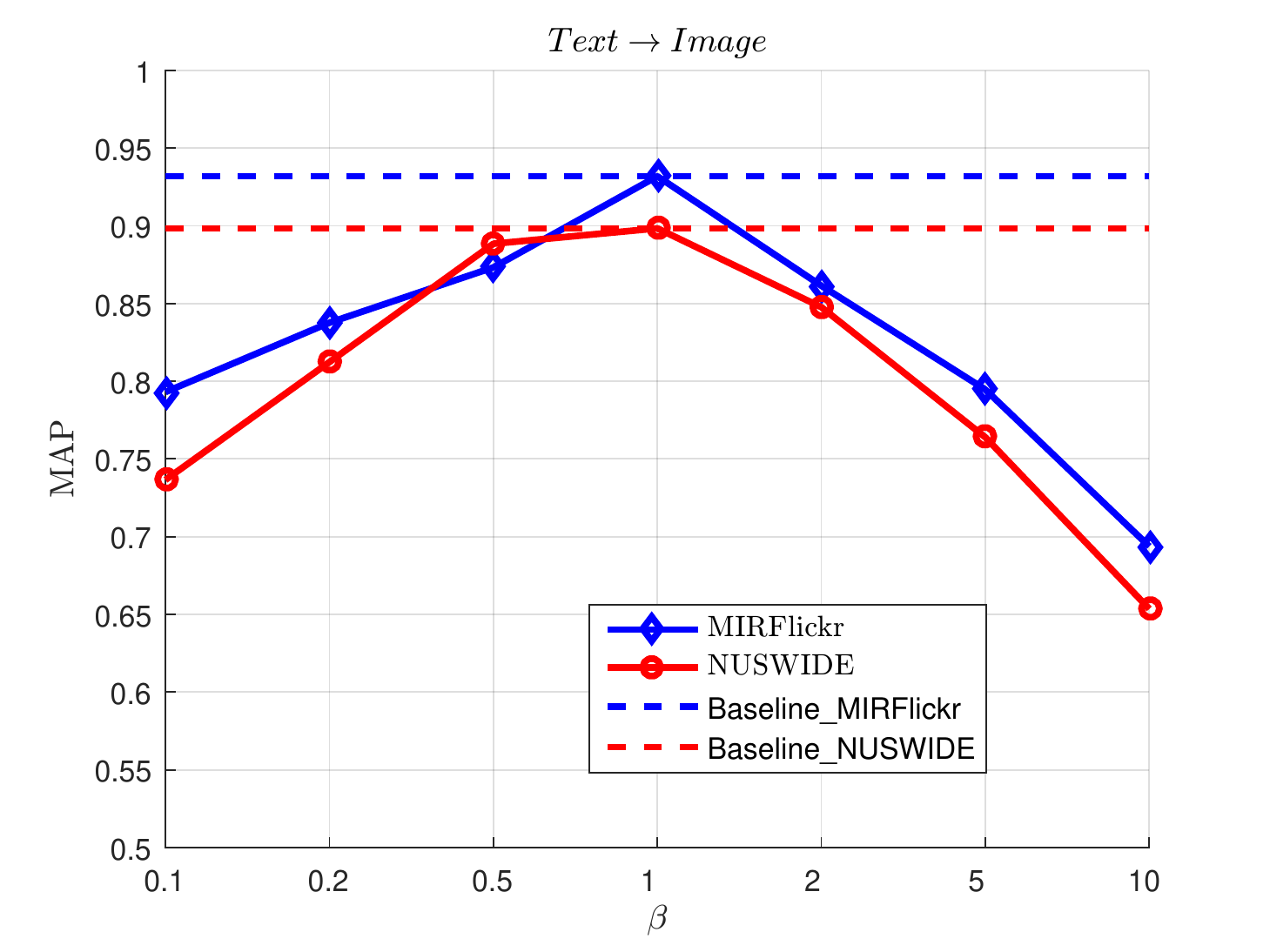}
		\end{minipage}
	}
	%\vspace{-0.3cm}
	\caption{MAP versus parameter $\beta$.}
	\label{fig: beta}
\end{figure}

The parameter $\beta$ leverages the importance of label alignment and quantization. When $\beta$ is large, quantization error will be small and approximate error in label alignment part will be large, and vice versa. To measure the concrete influence of $\beta$, we fix other parameters and report the MAP values by varying the value of $\beta$ between the range of $[0.1, 10]$. As shown in Fig.~\ref{fig: beta}, we can observe that the performance usually is improved with the increase of $\beta$ from $0.1$ to $1$. When $\beta$ is larger than $1$, the performance of SPDQ will decrease with the increase of $\beta$. So, in our experiments, we set $\beta$ to $1$.

\begin{figure}
	\centering
	\subfigure{
		\begin{minipage}[b]{0.98\textwidth}
			\includegraphics[width=0.24\textwidth]{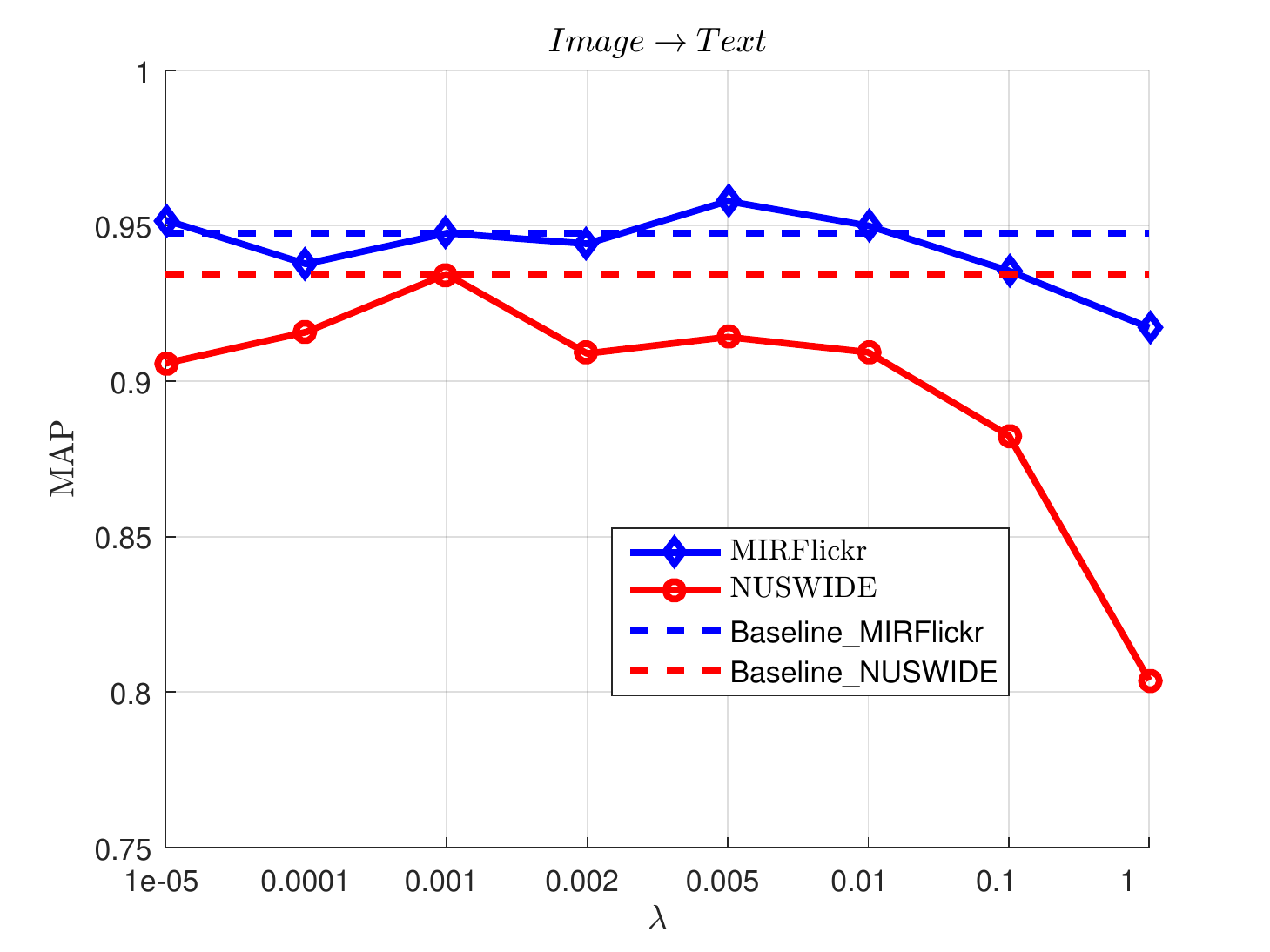}
			\includegraphics[width=0.24\textwidth]{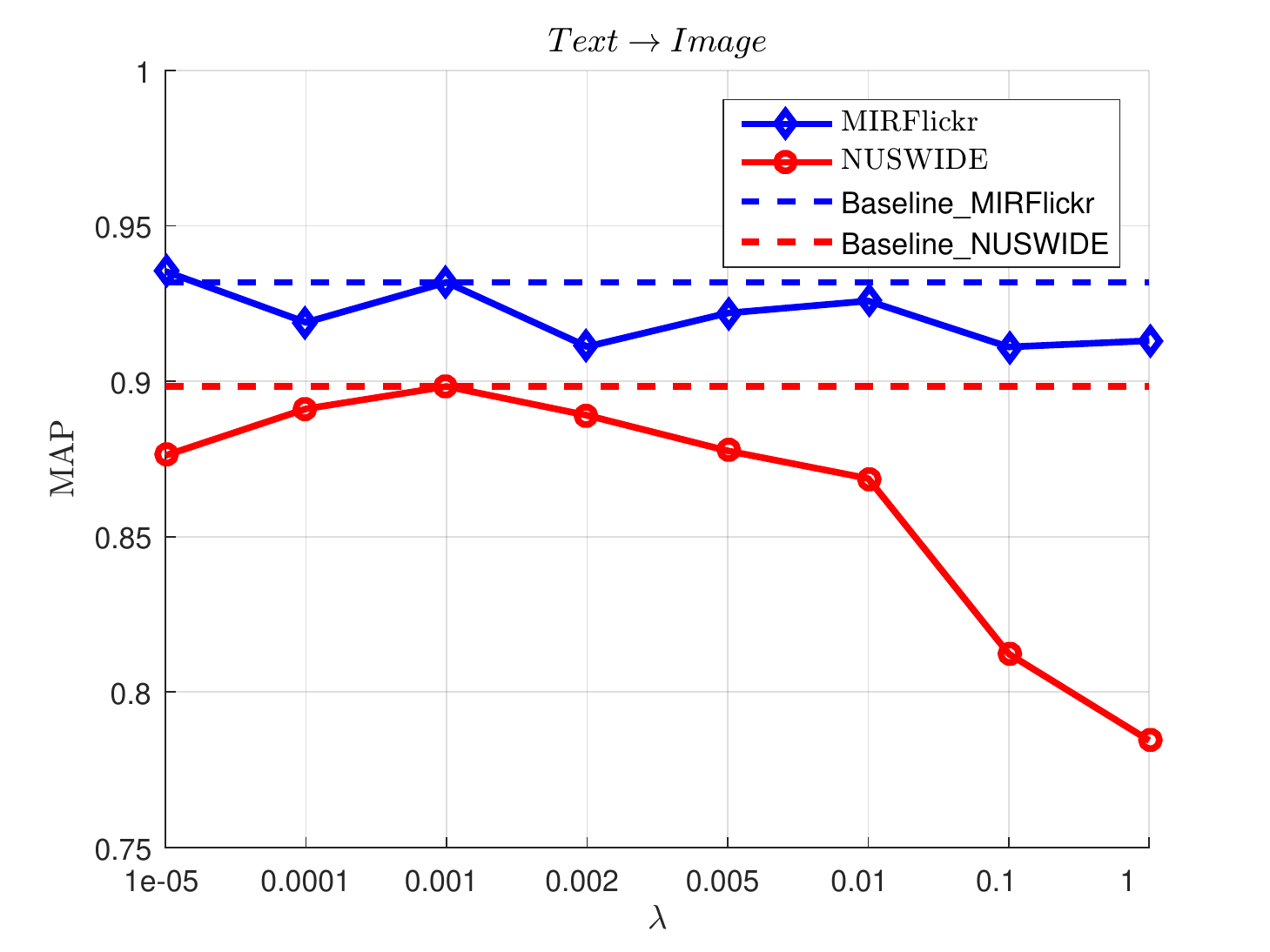}
		\end{minipage}
	}
	%\vspace{-0.3cm}
	\caption{MAP versus parameter $\gamma$.}
	\label{fig: gamma}
\end{figure}

The parameter $\gamma$ reflects the importance of the shared feature learning part and the label alignment quantization part. A proper value of $\gamma$ will enable SPDQ to learn shared features more suitable for quantization. To decide the optimal value of $\gamma$, we fix the hash code length as 32, and calculate the MAP values on FLICKR25K and NUS-WIDE dataset by varying $\gamma$ from $10^{-5}$ to $1$. The results are reported in Fig.~\ref{fig: gamma}. It can be observed that the performance on both datasets are relatively stable when $\gamma$ is small than $0.005$, and will decrease apparently when $\gamma$ is larger than $0.005$. Actually, we can select $\gamma$ from the range of $[10^{-5}, 5 \times 10^{-3}]$.

%-------------------------------------------------------------------------
\section{Conclusions}
\label{sec:conclusion}
In this paper, we proposed a novel quantization approach, namely shared predictive deep quantization (SPDQ), for efficient cross-modal similarity search. The superiority of the proposed approach lies in: 1) firstly exploiting a deep neural network to construct the shared subspace across different modalities and the private subspace for each modality, in which the correlations between multiple modalities can be well discovered and the specific characteristics of each modality can also be maintained; 2) introducing label alignment to the quantization training procedure,  thus preserving the semantic similarities of image-text pairs and greatly improving the search accuracy. The experimental results on two benchmark multi-modal datasets demonstrate that the proposed approach surpasses the existing methods.
\ifCLASSOPTIONcaptionsoff
  \newpage
\fi

% trigger a \newpage just before the given reference
% number - used to balance the columns on the last page
% adjust value as needed - may need to be readjusted if
% the document is modified later
%\IEEEtriggeratref{8}
% The "triggered" command can be changed if desired:
%\IEEEtriggercmd{\enlargethispage{-5in}}

% references section

% can use a bibliography generated by BibTeX as a .bbl file
% BibTeX documentation can be easily obtained at:
% http://mirror.ctan.org/biblio/bibtex/contrib/doc/
% The IEEEtran BibTeX style support page is at:
% http://www.michaelshell.org/tex/ieeetran/bibtex/
\bibliographystyle{IEEEtran}
% argument is your BibTeX string definitions and bibliography database(s)
\bibliography{SPDQ}
\end{document}